\documentclass[preprint,sort&compress,5p,twocolumn,lefttitle]{elsarticle}

\usepackage{dblfloatfix}

\usepackage{bm}
\usepackage{bbm}
\usepackage{gensymb}
\usepackage{graphicx}
\usepackage{amsmath}
\usepackage{amssymb}
\usepackage{algorithm}
\usepackage[noend]{algpseudocode}
\usepackage{subcaption}
\usepackage{amsfonts}
\usepackage[mode = match]{siunitx}
\usepackage{booktabs}
\usepackage{makecell}
\usepackage{multirow}
\usepackage{upgreek}
\usepackage{pifont}
\usepackage[font=small]{caption}
\usepackage[export]{adjustbox}
\usepackage{tikz}
\usepackage{sidecap} \sidecaptionvpos{figure}{c}
\usepackage{lscape}
\usepackage{threeparttable}

\DeclareMathOperator*{\argmax}{argmax}

\newcommand{\cmark}{\ding{51}}%
\newcommand{\xmark}{\ding{55}}%

\newcommand{\STAB}[1]{\begin{tabular}{@{}c@{}}#1\end{tabular}}
\newcommand\edit[1]{{\color{black}#1}}

\usepackage{hyperref}
\hypersetup{colorlinks,breaklinks,
linkcolor=[rgb]{0.5,0.,0.},
citecolor=[rgb]{0.000,0.427,0.173},
urlcolor=[rgb]{0.031,0.318,0.612}}

\usepackage[nameinlink]{cleveref}
\usepackage{color}
\definecolor{CommentPink}{rgb}{1,0.2,0.5}
\definecolor{CommentBlue}{rgb}{0,0,1}
\definecolor{CommentGreen}{rgb}{0,1,0}

\Crefname{section}{Sec.}{Sec.}
\Crefname{figure}{Fig.}{Fig.}
\Crefname{equation}{Eq.}{Eq.}

\usepackage[printonlyused,withpage,nolist,nohyperlinks]{acronym}
\begin{acronym}

\acro{2D}{two-dimensional}

\acro{3D}{three-dimensional}

\acro{AHRS}{attitude and heading reference system}

\acro{AUV}{autonomous underwater vehicle}

\acro{COMA}{counterfactual multi-agent policy gradients}

\acro{CPP}{Chinese Postman Problem}

\acro{DoF}{degree of freedom}
\acrodefplural{DoF}[DoFs]{degrees of freedom}

\acro{DVL}{Doppler velocity log}

\acro{FSM}{finite state machine}

\acro{IMU}{inertial measurement unit}

\acro{LBL}{Long Baseline}

\acro{MCM}{mine countermeasures}

\acro{MDP}{Markov decision process}
\acrodefplural{MDP}[MDPs]{Markov decision processes}

\acro{MCTS}{Monte Carlo Tree Search}

\acro{POMDP}{partially observable Markov decision process}
\acrodefplural{POMDP}[POMDPs]{partially observable Markov decision processes}

\acro{PRM}{Probabilistic Roadmap}
\acrodefplural{PRM}[PRM]{Probabilistic Roadmaps}

\acro{RL}{reinforcement learning}

\acro{ROI}{region of interest}
\acrodefplural{ROI}[ROIs]{regions of interest}

\acro{ROS}{Robot Operating System}

\acro{ROV}{remotely operated vehicle}

\acro{SLAM}{simultaneous localization and mapping}

\acro{SSE}{sum of squared errors}

\acro{STOMP}{Stochastic Trajectory Optimization for Motion Planning}

\acro{TRN}{Terrain-Relative Navigation}

\acro{UAV}{unmanned aerial vehicle}
\acrodefplural{UAV}[UAVs]{unmanned aerial vehicles}

\acro{UGV}{unmanned ground vehicle}
\acrodefplural{UAV}[UAVs]{unmanned ground vehicles}

\acro{USBL}{Ultra-Short Baseline}

\acro{IPP}{informative path planning}

\acro{AIPP}{adaptive informative path planning}

\acro{FoV}{field of view}
\acrodefplural{FoV}[FoVs]{fields of view}

\acro{CDF}{cumulative distribution function}

\acro{ML}{maximum likelihood}

\acro{RMSE}{Root Mean Squared Error}
\acro{MLL}{Mean Log Loss}

\acro{GP}{Gaussian Process}
\acrodefplural{GP}[GPs]{Gaussian Processes}

\acro{KF}{Kalman Filter}

\acro{IP}{Interior Point}
\acro{BO}{Bayesian Optimization}

\acro{SE}{squared exponential}

\acro{UI}{uncertain input}

\acro{MCL}{Monte Carlo Localisation}
\acro{AMCL}{Adaptive Monte Carlo Localisation}

\acro{SSIM}{structural similarity index measure}
\acro{PSNR}{peak signal-to-noise ratio}
\acro{LPIPS}{learned perceptual image patch similarity}

\acro{MAE}{Mean Absolute Error}
\acro{RMSE}{root mean squared error}
\acro{AUSE}{Area Under the Sparsification Error curve}
\acro{mIoU}{mean Intersection-over-Union}
\acro{CMA-ES}{covariance matrix adaptation evolution strategy}
\acro{RRT}{rapidly exploring random tree}
\acro{RIG}{rapidly exploring information gathering}

\acro{AL}{active learning}
\acro{MLP}{multi-layer perceptron}
\acro{LSTM}{long short-term memory}
\acro{PPO}{proximal policy optimization}
\acro{CNN}{convolutional neural network}

\acro{IL}{imitation learning}
\acro{GPS}{Global Positioning System}

\end{acronym}

\journal{Elsevier}

\begin{document}

\begin{frontmatter}



\title{\edit{Learning-based Methods} for \edit{Adaptive} Informative Path Planning}


\author[label1,label2]{Marija Popovi\'{c}\corref{cor1}} \ead{m.popovic@tudelft.nl}
\author[label3]{Joshua Ott} \ead{jott2@stanford.edu}
\author[label2]{Julius R\"{u}ckin} \ead{jrueckin@uni-bonn.de}
\author[label3]{Mykel J. Kochenderfer} \ead{mykel@stanford.edu}

\cortext[cor1]{Corresponding author.}

\affiliation[label1]{organization={MAVLab, Faculty of Aerospace Engineering, TU Delft},
            addressline={Kluyverweg 1}, 
            city={Delft},
            postcode={2629 HS}, 
            country={The Netherlands}}

\affiliation[label2]{organization={Cluster of Excellence PhenoRob, Institute of Geodesy and Geoinformation, University of Bonn},
            addressline={Niebuhrstr. 1A}, 
            city={Bonn},
            postcode={53113}, 
            country={Germany}}

\affiliation[label3]{organization={Stanford Intelligent Systems Laboratory, Department of Aeronautics and Astronautics, Stanford University},
            addressline={496 Lomita Mall}, 
            city={Stanford},
            postcode={94305}, 
            state={California},
            country={United States of America}}

\begin{abstract}
\Ac{AIPP} is important to many robotics applications, enabling mobile robots to efficiently collect useful data about initially unknown environments.
In addition, learning-based methods are increasingly used in robotics to enhance adaptability, versatility, and robustness across diverse and complex tasks.
Our survey explores research on applying robotic learning to \ac{AIPP}, bridging the gap between these two research fields.
We begin by providing a unified \edit{mathematical problem definition} for general \ac{AIPP} problems.
Next, we establish two complementary taxonomies of current work from the perspectives of (i) learning algorithms and (ii) robotic applications. We explore synergies, recent trends, and highlight the benefits of learning-based methods in \ac{AIPP} frameworks. 
Finally, we discuss key challenges and promising future directions to enable more generally applicable and robust robotic data-gathering systems through learning.
We provide a comprehensive catalog of papers reviewed in our survey, including publicly available repositories, to facilitate future studies in the field.

\end{abstract}



\end{frontmatter}

\section{Introduction}\label{S:introduction}
The field of robotics has witnessed remarkable advancements in recent years, fueled by the growing need for automation and the increasing complexity of tasks required in various application domains. One of the key challenges in robotics is \ac{AIPP}, which involves planning a trajectory for an autonomous robot to follow that maximizes the information acquired about an unknown environment while simultaneously respecting its resource constraints. This problem is of critical importance in applications such as environmental monitoring, exploration, search and rescue, and inspection across ground, aerial, and aquatic domains~\citep{rayas2022informative, cao2023catnipp, popovic2020informative, denniston2023fast, ruckin2022adaptive, viseras2019deepig, choudhury2020adaptive}.

Despite its importance, \ac{AIPP} is a challenging problem due to the inherent complexity of modeling and predicting new information in the environment, as well as the need to balance the exploration of unknown areas with the exploitation of newly acquired data~\citep{singh2009nonmyopic}.  Additionally, noisy sensor measurements introduce uncertainties into the data acquisition process, while actuation uncertainty can lead to deviations from planned trajectories. Moreover, real-world environments are often highly dynamic, which further complicates the modeling process and makes prediction challenging. Decisions must be made in a sequential manner to allow for continual refinement as more information becomes available during a mission. 

Conventional approaches such as static pre-computed paths tend to fail on \ac{AIPP} problems because they often rely on strong assumptions about the environment and cannot adapt to uncertainties or changes online~\citep{galceran2013survey}. Additionally, these approaches scale poorly to large, complex environments, and may not effectively account for the constraints and capabilities of the robot itself. These conventional solutions applied to the \ac{AIPP} problem have been limited by computational complexity, lack of adaptability, and an inability to generalize across diverse environments and problems~\citep{tan2021comprehensive}. The growing popularity of learning-based methods has led to a renewed focus on applying these new techniques to \ac{AIPP}, offering the promise of more flexible, adaptive, and scalable solutions.

\begin{figure*}[t]
    \centering
    \includegraphics[width=\textwidth]{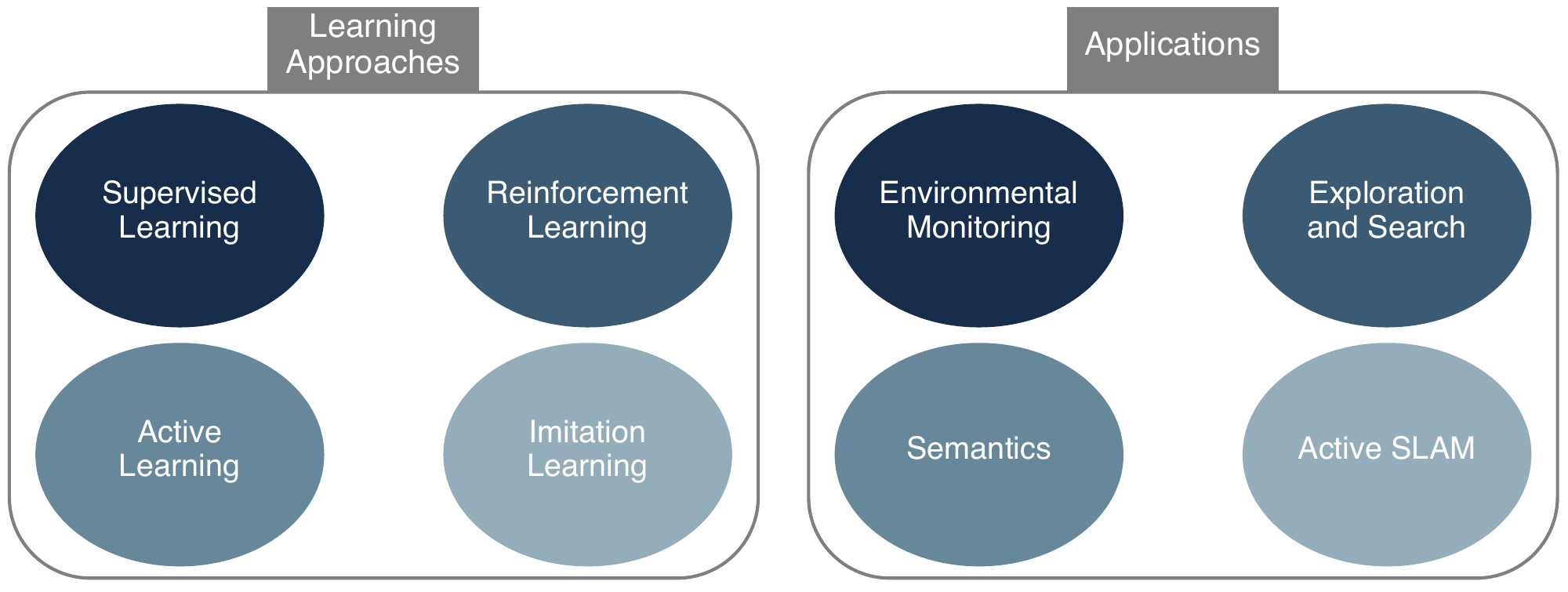}
\caption{Overview of our survey paper. We review related work in learning-based methods for adaptive informative path planning (AIPP) from the perspectives of learning-based approaches (left) and practical applications (right).} \label{F:overview}
\end{figure*}

In this survey paper, we present a new perspective on \ac{AIPP} by exploring the potential of emerging robot learning techniques in addressing the challenges of \ac{AIPP}. Different learning-based methods allow for constructing path planning algorithms that can naturally account for the inherent uncertainties and dynamic changes within a given environment. This adaptability facilitates more robust and efficient operations in complex environments because the robot can optimize its path based on its evolving understanding of its surroundings. This not only improves task performance, but also enhances resilience and versatility since the robot can react more effectively to new information about its environment.

\edit{\Cref{F:overview} shows an overview of the main topics addressed in our survey paper.} Our review includes different aspects of learning: supervised learning, reinforcement learning, imitation learning, and active learning. We focus on how these techniques can be used for \ac{AIPP} in robotics. Furthermore, we discuss relevant application domains, such as environmental monitoring, exploration and search, semantic scene understanding, and active \ac{SLAM}.

While there have been previous surveys addressing aspects of \ac{AIPP}~\citep{bai2021information,maboudi2023review,lluvia2021active,sung2023decision, taylor2021active, dos2022review}, our work distinguishes itself by specifically focusing on the intersection of learning-based methods and \ac{AIPP}. Previous literature reviews have focused on specific subfields of robotic learning. For instance, \citet{taylor2021active} focus on active learning in robotics, highlighting methods suitable for the demands of embodied learning systems. Similarly, \citet{argall2009survey} address imitation learning, where robots develop policies from example state-to-action mappings, while \citet{garaffa2023reinforcement} investigate reinforcement learning techniques that devise unknown environment \edit{exploration} strategies for single and multi-robot exploration. \Citet{lauri2022partially} exclusively examine the application of Partially Observable Markov Decision Processes (POMDPs) in robotics, and \citet{mukherjee2022survey} analyze robotic learning strategies for human-robot collaboration in industrial settings. Other surveys, such as those by \citet{sung2023decision}, \citet{dos2022review}, and \citet{bai2021information} explore different different applications of \ac{AIPP}, such as environmental monitoring tasks or general path planning techniques~\citep{bai2021information, galceran2013survey, chen2011kalman, tan2021comprehensive, maboudi2023review}. However, these works do not underscore the applicability of learning in adaptive scenarios.



The main objectives of this survey paper are to:
\begin{enumerate}
    \item provide a comprehensive understanding and taxonomy of the current state-of-the-art in learning-based methods for \ac{AIPP} as well as their applications;
    \item introduce a standardized \edit{mathematical problem definition for \ac{AIPP} tasks}, which provides a unified foundation for understanding and comparing various learning-based methods and application domains;
    \item identify potential avenues for future research by highlighting the limitations of current approaches.
\end{enumerate}

We have cataloged the papers in our survey at \url{https://dmar-bonn.github.io/aipp-survey/}. In particular, we highlight papers that have open-source implementations available.



\section{Mathematical Formulation}\label{S:formulation}
\edit{First, we develop an underlying mathematical formulation to encompass the works included in our survey.} We study the general problem of \acf{AIPP}. The goal is to find an optimal action sequence $\psi^* = (a_1, \ldots, a_N)$ of $N$ robot actions $a_i \in \mathcal{A}, i \in \{1, \ldots, N\},$ where actions may be, e.g., acceleration commands, next robot poses, using a specific sensor, or returning to a charging station. The action sequence $\psi^*$ maximizes an information-theoretic criterion $I(\cdot)$:
\begin{equation} \label{eq:ipp_problem}
    \psi^* = \argmax_{\psi \in \Psi} \, I(\psi),\, \text{s.t. } C(\psi) \leq B \, ,
\end{equation}
where $\Psi$ represents the set of all possible action sequences, the cost function $C: \Psi \to \mathbb{R}$ maps an action sequence to its associated execution cost, $B \in \mathbb{R}$ is the robot's budget limit, e.g., time or energy, and $I: \Psi \to \mathbb{R}$ is the information criterion, computed from the new sensor measurements obtained by executing the actions $\psi$.


Our problem setup considers a robot gathering sensor measurements in an initially or partially unknown environment. In this environment $\xi \subset \mathbb{R}^D$, \ac{AIPP} requires online replanning during a mission to \textit{adaptively} focus on areas of interest as they are discovered. Each point $\mathbf{x} \in \xi$, e.g., a $D$-dimensional pose within the environment, is characterized by its distinctive features, represented as \edit{$F(\mathbf{x},t) \in \mathcal{F}$}. Here, $\mathcal{F}$ is a feature space, which could include characteristics like the semantic class, temperature, or radiation level\edit{, and may vary over time}. These features may fluctuate during the mission, as defined by the feature mapping function  \edit{$F: \xi \times \mathbb{R} \to \mathcal{F}$}. The notation used to describe our \ac{AIPP} formulation is summarized in~\Cref{T:aipp_notation}.

Areas of interest within the environment the robot operates in are characterized by their features, which indicate interesting phenomena as defined by the mission objectives. Examples of mission objectives include identifying victims in a post-disaster scenario or locating regions of high radiation levels in a nuclear facility monitoring task. This concept of adaptivity in an optimal action sequence $\psi^*$ is reflected in the specific definition of the information-theoretic criterion $I$ during the optimization problem presented in~\Cref{eq:ipp_problem}.

Importantly, the point features \edit{$F(\mathbf{x},t)$}, such as temperature, may change over time. Given that the initial environment is unknown, the robot's understanding of \edit{$F(\mathbf{x},t)$} may need to be updated during the mission. This could result in updating beliefs regarding interesting regions over time as new sensor measurements are received. Thus, to ensure that the robot's behavior is adaptively informed, we may re-solve the constrained optimization problem in~\Cref{eq:ipp_problem} in a computationally efficient manner during a mission.

\begin{table*}[!t]
    \centering
    {\renewcommand{\arraystretch}{1.4}
    \begin{tabular}{@{}l ll@{}} 
     \toprule
     Symbol & Meaning & Examples \\
     \midrule
     $a_i \in \mathbb{A} $ & robot action & next robot position or pose, sensing behavior \\ 
     $\psi \in \Psi$ & sequence of consecutive actions & --- \\
     $\mathbf{x} \in \xi$ & areas/locations in an environment & surfaces, patches of terrain \\
     $I: \Psi \to \mathbb{R}$ & information-theoretic criterion & map entropy, mutual information \\
     \edit{$F: \xi \times \mathbb{R} \to \mathcal{F}$} & feature mapping function & semantic class, spatial occupancy, temperature, radiation level \\
     \bottomrule
    \end{tabular}}
    \caption{Notation associated with the adaptive informative path planning (AIPP) problem.\label{T:aipp_notation}}
\end{table*}

\section{Background}\label{S:background}
\begin{figure*}[!h]
    \centering
    \includegraphics[width=0.9\textwidth]{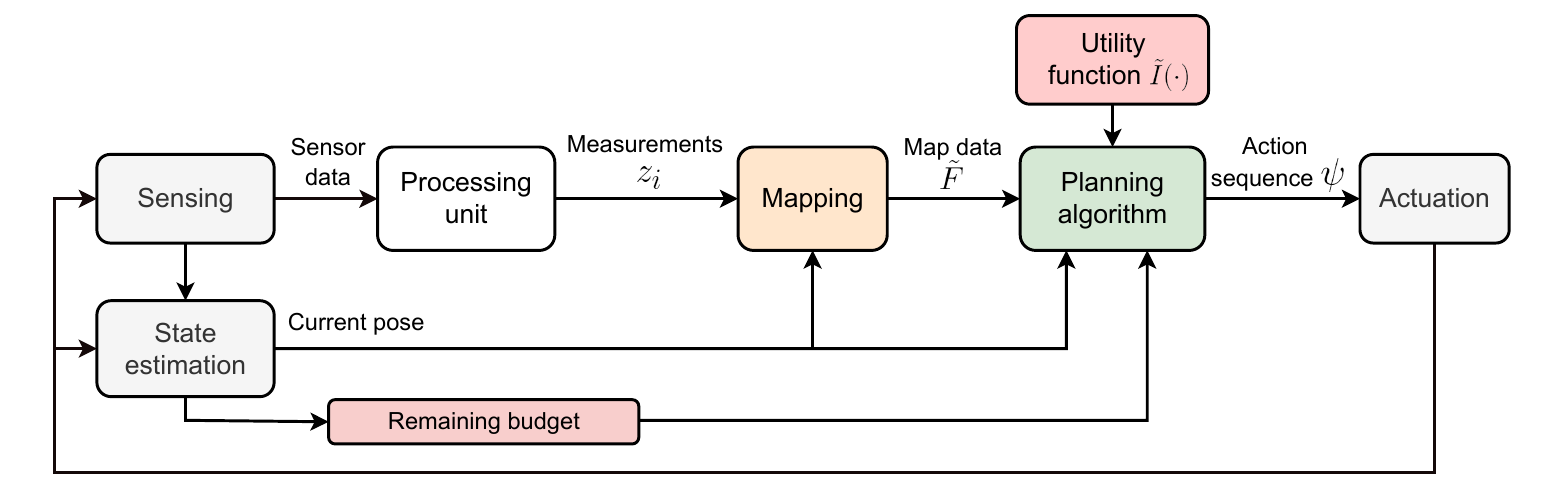}
    \caption{System diagram showing the key elements of a general \ac{AIPP} framework. During a mission, a map $\tilde{F}$ of the robot's environment $\xi$ is built using measurements $z$ extracted from a sensor data stream. An \ac{AIPP} algorithm leverages the map data to find the next action sequences $\psi$ for the robot to maximize the information value ${I}(\psi)$, i.e. utility, of future measurements. The next action sequence $\psi$ is executed by the robot, allowing for subsequent map updates in a closed-loop manner.}
    \label{F:ipp_system}
\end{figure*} 

The generic structure of a system for \ac{AIPP} is shown in~\Cref{F:ipp_system}. In an active sensing context, the framework processes raw data from a robot sensor to obtain measurements used to build a map of the robot's environment. During a mission, the \ac{AIPP} planning algorithm uses the environment maps built online to optimize future action sequences for maximum gain in an information metric as defined by the utility function. This section describes methods for environment mapping, evaluation metrics, and standard benchmarks commonly used in \ac{AIPP} frameworks.

\subsection{Mapping} \label{SS:mapping}

The \ac{AIPP} problem assumes the environment $\xi$ to be \textit{a priori} (partially) unknown. Thus, the true feature mapping \edit{$F: \xi \times \mathbb{R} \to \mathcal{F}$} and associated true regions of interest are \textit{a priori} (partially) unknown as well. This makes the original \ac{AIPP} problem stated in \Cref{eq:ipp_problem} not only computationally challenging but also impossible to solve directly as exact knowledge about interesting regions requires ground truth access to the feature mapping $F$. Instead of solving \Cref{eq:ipp_problem} directly, we model a stochastic process $\tilde{F}$ over all possible feature mapping functions. At each mission time step $t$, the robot's belief about the true feature mapping $F$, i.e., the probability measure associated with $\tilde{F}$ conditioned on the action-observation history, is updated by:
\begin{equation} \label{eq:map_belief}
    p(F \mid z_1, \ldots, z_t, a_1, \ldots, a_{t}) \, ,
\end{equation}
where $\{z_1, \ldots, z_t\}$ and $\{a_1, \ldots, a_{t}\}$ are all previously collected measurements and executed actions, respectively. The belief about $\tilde{F}$ is used to approximate the true information criterion $I$ in~\Cref{eq:ipp_problem}. 

The robot's belief about $\tilde{F}$ in~\Cref{eq:map_belief} may be updated online during a mission as new sensor data $z_t$ arrives at time step $t$ after executing a sensing action $a_t$. \edit{Note that the information criterion $I(\cdot)$ reflects the planning objective and is computed based on the map state. In general, the computation of certain information criteria, e.g. entropy, may change for different map representations, e.g. occupancy maps and Gaussian processes, but the key underlying theoretical concepts remain the same.}

The measurements $z_t$ obtained at time step $t$ may be uncertain or noisy. They are assumed to be sampled from $p(z \,|\, F)$ representing a probabilistic sensor model. Furthermore, there may be uncertainty in localization and actuation, leading to imperfect estimation of the robot pose as measurements are taken. In~\Cref{S:learning_approaches} and~\Cref{S:applications}, we discuss in detail how these different sources of uncertainty can be accounted for in the context of learning-based methods.

To compute the belief $\tilde{F}$ following~\Cref{eq:map_belief} based on the collection of stochastic measurements $\{z_1, \ldots, z_t$\}, various methods for \textit{environment mapping} are commonly used. \Cref{T:mapping_metrics} provides an overview of commonly used mapping methods and corresponding metrics derived from them used to evaluate the performance of \ac{AIPP} methods. We consider four broad categories in our taxonomy: occupancy grid mapping~\citep{elfes1989using}, Gaussian processes~\citep{rasmussen2006gaussian}, graph-based methods, and implicit neural representations~\citep{mildenhall2021nerf}. In general, the chosen mapping method for a given \ac{AIPP} scenario depends on the environment, task, and available computing resources. For instance, Gaussian processes are applicable for problems where the to-be-mapped features \edit{$F(\mathbf{x},t)$} support continuous feature spaces $\mathcal{F}$ and the map supports continuous environment representations $\xi$, whereas occupancy grid mapping is suitable for discrete mapping tasks. In contrast, graph-based methods are typically used for active \ac{SLAM} problems~\citep{placed2023survey} where both environment mapping and robot localization are concurrent. Recently, implicit neural representations, such as NeRFs~\citep{mildenhall2021nerf} and pixelNeRFs~\citep{yu2020pixelnerf}, are gaining popularity for \ac{AIPP} without relying on an explicit environment map. Due to their lightweight nature, these methods enable conserving computational and memory resources during planning in 3D reconstruction tasks where a high level of detail is required.

\begin{table}[!h]
  \centering
  \begin{minipage}{.99\columnwidth}
  {\renewcommand{\arraystretch}{1.15}
  \setlength{\tabcolsep}{4pt}
  \scriptsize
    
  \begin{tabular}{@{}c l l@{}}
    \toprule
     & \multicolumn{1}{l}{Evaluation metrics} & \multicolumn{1}{l}{References} \\ \midrule
     
     \multirow{11}{*}{\STAB{\rotatebox[origin=c]{90}{Occupancy grid}}}
     & Entropy      & \citep{popovic2020informative,westheider2023multi,bai2017toward,chen2020autonomous,sukkar2019multi,choudhury2018data}     \\
     & Mutual information       & \citep{lodel2022where,chen2019self} \\
     & Root mean squared error       & \citep{popovic2020informative} \\
     & Mean absolute error       & \citep{chen2021zeroshot,viseras2021wildfire} \\
     & Mean intersection-over-union      & \citep{georgakis2022upen,ramakrishnan2020occupancy}    \\
     & F1-score & \citep{westheider2023multi} \\
     & Accuracy & \citep{schmid2022scexplorer,georgakis2022upen,ramakrishnan2020occupancy} \\
     & Precision/recall      &  \citep{schmid2022scexplorer}   \\
     & Coverage      &  \citep{schmid2022scexplorer,schmid2022fast,niroui2019deep,cao2023adriadne,reinhart2020learning,zeng2022deep,shrestha2019learned,chen2021zeroshot,zacchini2023informed,liu2023learning,vasquez2021next,mendoza2020supervised,caley2019deep,georgakis2022upen,ramakrishnan2020occupancy,chen2022learning,tao2023seer,li2023learning,gao2022cooperative,sukkar2019multi,zwecher2022integrating,dai2022camera} \\ 
     & Observed surfaces      &  \citep{hepp2018learn,dhami2023pred,pan2023oneshot}  \\
     & Landmark error      &  \citep{choudhury2020adaptive}   \\
     \midrule
     
     \multirow{10}{*}{\STAB{\rotatebox[origin=c]{90}{Gaussian process}}}
     & Entropy     &  \citep{denniston2023learned}  \\
     & Expected improvement     &  \citep{denniston2023learned,song2015trajectory}  \\
     & Probability of improvement     &  \citep{denniston2023learned}  \\
     & Covariance matrix trace     &  \citep{cao2023catnipp,ruckin2022adaptive,popovic2020informative,popovic2020informativelocunc,ott2023sboaippms,yang2023intent,yanes2023deep}  \\
     & Mutual information     &  \citep{hitz2017adaptive,wei2020informative}     \\
     & Root mean squared error       & \citep{cao2023catnipp,ruckin2022adaptive,rayas2022informative,vivaldini2019uav,popovic2020informative,popovic2020informativelocunc,hollinger2014sampling,viseras2019deepig,yanes2023deep,choi2021adaptive}   \\
     & Mean log likelihood      &  \citep{popovic2020informative}    \\
     & Distribution divergence      &  \citep{wang2023spatiotemporal}    \\
     & Mean uncertainty       & \citep{wang2023spatiotemporal,hollinger2014sampling}       \\
     & Upper confidence bound       & \citep{marchant2014sequential}  \\ \midrule
     
    \multirow{7}{*}{\STAB{\rotatebox[origin=c]{90}{Graph-based}}}
     & Landmark uncertainty     &  \citep{chen2021zeroshot, chen2020autonomous}  \\
     & Landmark error     &  \citep{chen2020autonomous}     \\
     & Localization uncertainty     &  \citep{chen2020autonomous, popovic2020informativelocunc}  \\
     & Localization error       & \citep{popovic2020informativelocunc, huttenrauch2019deep} \\
     & Target uncertainty  & \citep{duecker2021embedded,tzes2023graph, singh2009nonmyopic} \\
     & Detection rate  & \citep{best2019dec} \\
     & Path costs & \citep{meliou2007nonmyopic} \\
     \midrule
     
    \multirow{11}{*}{\STAB{\rotatebox[origin=c]{90}{Neural representation}}}
     & Peak signal-to-noise ratio     &  \citep{jin2023neunbv, pan2022activenerf, ran2022neurar, zhan2022activermap, sunderhauf2023density, pan2023views}  \\
     & Structural similarity index measure     &  \citep{jin2023neunbv, pan2022activenerf, ran2022neurar, zhan2022activermap, pan2023views}  \\
     & Learned perceptual image patch similarity     &  \citep{pan2022activenerf,ran2022neurar, zhan2022activermap}  \\
     & Accuracy     &  \citep{ran2022neurar, zhan2022activermap,yan2023active}  \\
     & Completion     &  \citep{ran2022neurar, zhan2022activermap,yan2023active}  \\
     & Completion ratio     &  \citep{ran2022neurar,yan2023active}  \\
     & F1-score     &  \citep{zhan2022activermap, lee2022uncertainty}  \\
     & Chamfer distance     &  \citep{zhan2022activermap}  \\
     & Precision     &  \citep{lee2022uncertainty}  \\
     & Recall     &  \citep{lee2022uncertainty}  \\
     & Surface coverage     &  \citep{lee2022uncertainty,yan2023active}  \\ \bottomrule     
  \end{tabular}}
  \caption{Mapping methods and evaluation metrics used in \ac{AIPP}.\label{T:mapping_metrics}}
  \end{minipage}
  
\end{table}

\subsection{Evaluation Metrics} \label{SS:evaluation}

The goal of \ac{AIPP} is to adaptively acquire information about initially unknown environment features $F$ by collecting uncertain or noisy measurements $z_t$ during a mission, usually captured as a posterior map belief $\tilde{F}$ in~\Cref{eq:map_belief}. Because the robot mission goals are task- and application-dependent, e.g., searching for targets~\citep{ott2023sboaippms,tzes2023graph,wang2023spatiotemporal} or regions of high surface temperature~\citep{ruckin2022adaptive,meliou2007nonmyopic,westheider2023multi}, there is a lack of standard evaluation metrics used to quantify the performance of \ac{AIPP} algorithms. Most works choose evaluation metrics closely related to their specified information criterion $I$ in~\Cref{eq:ipp_problem} because they reflect or approximate the mission goal. \Cref{T:mapping_metrics} lists commonly used evaluation metrics. Although these metrics vary significantly in the \ac{AIPP} community, most works consider quantitative measures based on the map representation to compute the belief $\tilde{F}$, as reflected by our taxonomy. 

Applications aiming to collect information about discrete features $F$, such as spatial occupancy~\citep{bai2017toward,chen2020autonomous} or land cover classification~\cite{popovic2020informative,vivaldini2019uav,rayas2022informative}, commonly use the entropy of the belief $\tilde{F}$ as a measure for the remaining uncertainty of the environment features~\citep{popovic2020informative,westheider2023multi,bai2017toward,chen2020autonomous,sukkar2019multi,choudhury2018data}. Lower entropy of $\tilde{F}$ indicates better \ac{AIPP} performance. Following popular computer vision benchmarks~\cite{everingham2010pascal, cordts2016cityscapes}, some works evaluate the classification quality of $\tilde{F}$ by computing the \ac{mIoU}~\citep{georgakis2022upen,ramakrishnan2020occupancy}, accuracy~\citep{schmid2022scexplorer,georgakis2022upen,ramakrishnan2020occupancy}, or F1-score~\citep{westheider2023multi} of the maximum a posteriori estimate of the map belief $\tilde{F}$ given the ground truth features $F$.

Methods gathering information about continuous features $F$ in an environment, e.g., bacteria level in a lake~\citep{hitz2017adaptive} or magnetic field strength~\citep{viseras2019deepig}, often use Gaussian process models~\cite{hitz2017adaptive, cao2023catnipp, viseras2019deepig, ott2023sboaippms} or some variant of Bayesian filtering~\cite{popovic2020informative, ruckin2022adaptive, choudhury2020adaptive, denniston2023fast} as the map representations. Although Gaussian processes allow for computing the differential entropy of $\tilde{F}$ in closed-form~\citep{williams2006gaussian}, it is common to use the covariance matrix trace to approximate the remaining uncertainty about the environment features and hence the information criterion $I$ due to its computational efficiency. Despite relaxed computational requirements, most works evaluate \ac{AIPP} performance based on the covariance matrix trace during offline evaluation as well~\citep{cao2023catnipp,ruckin2022adaptive,popovic2020informative,popovic2020informativelocunc,ott2023sboaippms,yang2023intent,yanes2023deep}. To evaluate the reconstruction quality of $\tilde{F}$, the \ac{RMSE} between the maximum \textit{a posteriori} estimate of $\tilde{F}$ and the ground truth $F$ is usually reported~\cite{cao2023catnipp,ruckin2022adaptive,rayas2022informative,vivaldini2019uav,popovic2020informative,popovic2020informativelocunc,hollinger2014sampling,viseras2019deepig,yanes2023deep,choi2021adaptive}.

Evaluation methods for implicit neural representations differ in that they only rely on a trained neural network to represent the entire environment. The metrics can be grouped based on two aspects: the quality of the rendered images and the quality of the geometry of the reconstructed surface. Image quality metrics, e.g., \ac{PSNR}, \ac{SSIM}, and \ac{LPIPS}, are computed based on a set of evenly distributed test images in the environment~\citep{jin2023neunbv,pan2022activenerf,sunderhauf2023density,ran2022neurar,zhan2022activermap,pan2023views}. To evaluate the reconstruction quality, the learned geometric representation is usually first post-processed, e.g., by using the marching cubes algorithm to extract a mesh. Then, a set of points is sampled on the mesh to compute various metrics, e.g., mesh accuracy or completion~\citep{lee2022uncertainty,ran2022neurar,zhan2022activermap}.

Some methods do not assume the robot to have perfect ground truth localization information during missions but additionally account for localization uncertainty while building the map belief $\tilde{F}$. These works propagate localization uncertainty into the map belief update in~\Cref{eq:map_belief} using graph-based \ac{SLAM} systems~\citep{chen2020autonomous, chen2021zeroshot, popovic2020informativelocunc,tzes2023graph} and Gaussian process variants supporting input uncertainty~\citep{popovic2020informativelocunc}. In addition to assessing the quality of the map, the quality of robot localization is also evaluated to determine whether the \ac{AIPP} method can gather not only informative measurements to improve the environment map, but also measurements that improve localization to guarantee the construction of an accurate map. Some works compute the trajectory errors between the estimated and ground truth executed trajectory~\cite{chen2021zeroshot}. Other works track the localization uncertainty of the robot pose belief~\cite{popovic2020informativelocunc, chen2020autonomous}, the localization error between the estimated and ground truth robot pose~\cite{popovic2020informativelocunc,huttenrauch2019deep}, or the landmark errors between estimated and ground truth landmark poses~\citep{chen2020autonomous}.

An alternative problem formulation considers using \ac{AIPP} to actively improve the sensor model reasoning about measurements $z_t$. In~\Cref{eq:ipp_problem}, this is done by linking the information criterion $I$ to the value of new training data acquired during a mission. Such formulations generally do not aim to improve the belief of the map $\tilde{F}$ representing a physically quantifiable environmental phenomenon. Instead, works in this category target improving a neural network for semantic segmentation using RGB images~\citep{blum2019active, zurbrugg2022embodied, ruckin2023informative, ruckin2023semi} or object detection~\citep{chaplot2021seal}. \ac{AIPP} performance is evaluated using a held-out test set to quantify the neural network prediction performance~\cite{everingham2010pascal, cordts2016cityscapes}. Commonly used metrics include \ac{mIoU}~\cite{blum2019active, zurbrugg2022embodied, ruckin2023informative}, F1-score~\citep{ruckin2023informative,ruckin2023semi}, or average precision~\citep{chaplot2021seal}, depending on the prediction task.

Due to the sequential nature of the \ac{AIPP} problem, most works consider a well-performing \ac{AIPP} approach not only to show strong evaluation metrics after a mission is over but also to ensure fast improvement during a mission as the remaining mission budget $B$ decreases. Trends in performance over mission time allow us to judge the efficiency of budget allocation, which is critical in \ac{AIPP} as reflected by the constraints in~\Cref{eq:ipp_problem}. Common evaluation strategies include analyzing how the performance-based evaluation metrics evolve with the remaining mission budget, e.g., trajectory steps~\citep{ott2023sboaippms, westheider2023multi, viseras2019deepig, cao2023catnipp, denniston2023fast} or energy-based budgets~\citep{popovic2020informative, ruckin2022adaptive, hollinger2014sampling, choudhury2020adaptive}.

\subsection{Benchmarks} \label{SS:benchmarks}

Despite research efforts towards more generally applicable learning-based \ac{AIPP}, most current methods are evaluated in application-dependent simulators and compared to task-dependent baseline algorithms. This results in a lack of standardized benchmark scenarios within the \ac{AIPP} community. Benchmark scenarios are typically defined by two factors: (i) the choice of simulation environments and (ii) the choice of \ac{AIPP} baseline algorithms against which the proposed methods are compared. Simulators fulfill a role analogous to datasets in the computer vision community as they define the problem setup in terms of the mission objective, evaluation environment, and robot configuration. \ac{AIPP} baseline algorithms can be seen as paralleling widely established computer vision models, which are typically used to benchmark new models according to evaluation metrics.

\edit{\subsubsection{Simulations}}

Simulators for \ac{AIPP} evaluation make various interdependent design choices leading to highly specific simulation environments. First, \ac{AIPP} methods are typically evaluated on a single fixed task, e.g., the RockSample task~\citep{ott2023sboaippms, choudhury2020adaptive}, monitoring missions~\cite{popovic2020informative, ruckin2022adaptive, westheider2023multi, cao2023catnipp,choi2021adaptive,hollinger2014sampling}, or reconstruction~\citep{zeng2022deep, maboudi2023review,mendoza2020supervised,gazani2023bag,dhami2023pred,ran2022neurar, zhan2022activermap,yan2023active}. Second, simulators differ in the robot configuration defined by the robot type, e.g., \ac{AUV}~\citep{duecker2021embedded}, \ac{UAV}~\citep{popovic2020informative}, \ac{UGV}~\citep{wei2020informative}, or a robot arm~\citep{zeng2022deep}, as well as the sensors, e.g., depth sensors~\citep{cao2023adriadne,liu2023learning,niroui2019deep,hepp2018learn}, thermal camera~\citep{ruckin2022adaptive}, point-based sensors~\cite{cao2023catnipp, hitz2017adaptive,yang2023intent,marchant2014sequential}, or abstractions of sensor models interpreting raw sensor data~\citep{popovic2020informative, chaplot2021seal, denniston2023fast, georgakis2022learning,denniston2023learned}. Third, works vary in their assumptions about the structure of environments, e.g., 2D terrains~\citep{hitz2017adaptive, popovic2020informative, cao2023catnipp, ruckin2022adaptive, ott2023sboaippms, choudhury2020adaptive} or 3D volumes~\citep{cao2023adriadne,zeng2022deep, rayas2022informative}, and robot workspaces, e.g., obstacle-free 2D~\citep{cao2023catnipp,velasco2020adaptive}, obstacle-free 3D~\citep{popovic2020informative, westheider2023multi, ruckin2022adaptive,denniston2023learned,rayas2022informative}, or obstacle-aware 3D~\citep{cao2023adriadne,zeng2022deep,hepp2018learn} workspaces.  Last, simulators vary according to the datasets they use, e.g., synthetic environments or randomly generated distributions~\citep{popovic2020informative, cao2023catnipp, hitz2017adaptive, ruckin2022adaptive, westheider2023multi,schmid2022fast,bai2017toward,saroya2020online}, more realistic simulators~\citep{zeng2022deep,ruckin2023informative,zacchini2023informed,zhang2023affordance}, or real-world remote sensing datasets~\cite{popovic2020informative, ruckin2023informative,westheider2023multi, rayas2022informative,hitz2017adaptive, hollinger2014sampling,caley2019deep}. \Cref{S:learning_approaches} and \Cref{S:applications} provide a comprehensive overview of these aspects for each paper studied in our survey. In general, diverse non-unified simulator choices result in task- and application-dependent evaluation protocols. As a result, it is difficult to compare the performance of different \ac{AIPP} algorithms in a generalizable and fair fashion under consistent conditions.

\edit{\subsubsection{Common and Classical Baselines}}

Despite the lack of standardized \ac{AIPP} algorithms used for benchmarking new methods, some simple baseline algorithms have been established and used across different tasks and applications. These algorithms aim to foster exploration of the environment and use intuitive hand-crafted heuristics to optimize the evaluation metrics. We categorize approaches for exploration into two groups:
\begin{enumerate}
    \item Geometric approaches, which pre-compute paths to maximize coverage of the initially unknown environment for uniform data collection. Examples include lawnmower-like grid patterns~\citep{blum2019active,westheider2023multi,galceran2013survey} or spiral- or circle-like patterns~\cite{zeng2022deep, popovic2020informative,zhang2023affordance}.
    \item  Random walk-like methods, which sample paths at random often combined with heuristics to efficiently manage the mission budget~\citep{zurbrugg2022embodied,dai2022camera,kumar2022graph,sukkar2019multi,liu2023learning}.
\end{enumerate}

In addition to these simple non-adaptive benchmarks, many works consider variants of well-established planning algorithms in the \ac{AIPP} context. Classical planners used for evaluation include branch-and-bound techniques~\citep{binney2012branch} and sampling-based methods, e.g., \ac{MCTS}~\cite{choudhury2020adaptive}, \ac{RIG} trees~\cite{hollinger2014sampling}, or \acp{RRT}~\citep{karaman2011sampling}. Other benchmarks include geometric strategies, e.g., greedy frontier-based exploration~\citep{yamauchi1997frontier}, or optimization-based routines, e.g., Bayesian optimization~\citep{gelbart2014bayesian} or the \ac{CMA-ES}~\citep{hansen2001completely}. To enable adaptive replanning during a mission as new sensor measurements arrive, these algorithms are commonly implemented in a fixed- or receding-horizon fashion.

\section{Learning Approaches}\label{S:learning_approaches}

This section surveys the \ac{AIPP} literature from a learning-based perspective. We provide a taxonomy classifying relevant works based on the underlying learning algorithm they use, discuss pertinent aspects in each category, and integrate them in our \edit{mathematical problem definition} for \ac{AIPP}.

\subsection{Supervised Learning} \label{SS:supervised_learning}
Traditionally, the components of the general \ac{AIPP} framework in~\Cref{F:ipp_system} are realized using hand-crafted models or heuristics, e.g., forward sensor models~\citep{popovic2020informative,lim2016adaptive,karaman2011sampling,ruckin2022adaptive} or hand-crafted information criteria~\citep{hollinger2014sampling,binney2012branch,best2019dec,isler2016information}. This makes it difficult to incorporate prior knowledge and the expected distributions of geometry and information in complex scenarios, as well as to achieve adaptability in new environments. To address these drawbacks, various methods have been proposed to learn different elements of mapping and planning through supervised training from labeled data.

Given a dataset of $N$ training examples of the form $\{(x_1, y_1), \ldots, (x_N,y_N)\}$, a supervised learning algorithm seeks to learn a ground truth function $g: X \rightarrow Y$, where $X$ and $Y$ are the input and output spaces, respectively, and $(x_i, y_i) \in X \times Y$ are dataset pairs. \Cref{T:supervised_learning} provides a taxonomy of supervised learning applications in \ac{AIPP}. These methods find diverse usage in both mapping and planning.

Supervised learning has been used for modeling environmental variables through Gaussian processes~\citep{rayas2022informative,cao2023catnipp,ruckin2022adaptive,yang2023intent,wang2023spatiotemporal,marchant2014sequential,hitz2017adaptive,popovic2020informative,hollinger2014sampling,vivaldini2019uav,viseras2019deepig,yanes2023deep,choi2021adaptive,wei2020informative,song2015trajectory,denniston2023learned}. As described in~\Cref{SS:mapping}, Gaussian processes are often used in~\Cref{eq:map_belief} due to their probabilistic and non-parametric nature, making them suitable for uncertainty-based information gathering with spatially complex underlying feature mappings $F$. In this context, supervised learning is used to compute the ground truth function $g = F$ from a training dataset $(x_i, z_i) \in \xi \times \mathcal{F}$, with $X = \xi$ and $Y = \mathcal{F}$, obtained by taking stochastic measurements $z_i$ in the environment. Note that we use $x_i$ to denote general training inputs and $\mathbf{x}_i$ to denote the robot pose. In Gaussian process mapping, each $x_i$ corresponds to the pose $\mathbf{x}_i \in \xi$ introduced in~\Cref{S:formulation}. Example applications include mapping distributions of temperature~\citep{popovic2020informativelocunc,westheider2023multi,ruckin2022adaptive}, light~\citep{denniston2023fast,cao2023catnipp}, water properties~\citep{hitz2017adaptive,yanes2023deep}, and vegetation~\citep{popovic2020informative,vivaldini2019uav}. The learning process in implicit neural representations can be interpreted similarly. These methods learn a function that maps 3D coordinates $X = \xi$ to radiance values $Y = \mathcal{F}$, which represent color and light intensity, from a dataset of posed RGB~\citep{jin2023neunbv,pan2022activenerf,zhan2022activermap,sunderhauf2023density,pan2023views} or RGB\nobreakdash-D~\citep{ran2022neurar,yan2023active,lee2022uncertainty,zhang2023affordance} images.

For robotic exploration, several works focus on learning map completion from a partially observed map state, represented by its geometry~\citep{shrestha2019learned,zwecher2022integrating,georgakis2022upen,tao2023seer,dhami2023pred,li2023learning}, topology~\citep{saroya2020online}, or semantics~\citep{schmid2022scexplorer}. At mission time step $t$, these methods predict the evolution of a probabilistic map representation $k$ steps into the future $\tilde{F}_{t+k}$. The dataset pairs $(x_i, y_i)$ used for learning map completion are application-dependent; for example, learning inputs can range from local observations~\citep{schmid2022scexplorer} to the current global map state~\citep{li2020graph}. The key advantage of such approaches is their ability to extrapolate incomplete maps by minimizing a specific loss during training. This extrapolation enables better reasoning with limited information and occlusions compared to classical heuristic strategies that plan solely based on the map state at a given time step.

Several works have also investigate the benefits of using supervised learning in the planning algorithm. One straightforward approach is to learn the information criterion $I(\cdot)$ in~\Cref{eq:ipp_problem} from a dataset of actions and their associated information gains observed in previous data gathering missions $(x_i,y_i) = (a_i, I(a_i))$, where $X = \mathcal{A}$ and $Y = \mathbb{R}$~\citep{hepp2018learn,bai2017toward,ly2019autonomous}. To solve the \ac{AIPP} problem in~\Cref{eq:ipp_problem} directly, the learning process can depend on pairs that connect map states to corresponding actions. Each action $a_i$ is created using a planning algorithm with full access to the ground truth data represented by $g$.
\citet{schmid2022fast} and \citet{zacchini2023informed} propose learning a probabilistic model of informative views in the context of sampling-based planning. Alternatively, the next best viewpoint can be learned by reasoning about the shape of an object to be reconstructed~\citep{vasquez2021next,mendoza2020supervised} or localization performance in the context of active \ac{SLAM}~\citep{hanlon2023active}. When the dataset pairs $(x_i, y_i)$ are generated through expert demonstrations, supervised learning can be thought of as a simple form of imitation learning (IL) as described in~\Cref{SS:imitation_learning}. By learning the information criterion $I(\cdot)$ from data, these approaches all bypass formulating it explicitly for online replanning.


A key requirement for all methods in~\Cref{T:supervised_learning} is the availability of reliable training data for model supervision. Some supervised learning methods in \ac{AIPP} rely on open-source datasets, such as the indoor Matterport3D dataset~\citep{georgakis2022upen,tao2023seer,li2023learning} or outdoor 3D Street View dataset~\citep{hepp2018learn}. Works in environmental monitoring may use data from previously executed missions~\citep{rayas2022informative,hitz2017adaptive,hollinger2014sampling,caley2019deep}, while others create custom synthetic environments approximating real-world scenarios~\citep{schmid2022fast,bai2017toward,saroya2020online,cao2023catnipp,ruckin2022adaptive,yang2023intent,hitz2017adaptive,popovic2020informative,popovic2020informativelocunc,viseras2019deepig,zwecher2022integrating,song2015trajectory,marchant2014sequential}. However, the lack of realistic labeled training data for \ac{AIPP} problems remains an open issue restricting the generalizability of existing methods to new environments and domains. Furthermore, since supervised learning models are trained only on static data, their applicability is limited in dynamic environments with changing obstacles or terrains or the target information distribution varies over time. Finally, as most deep learning models do not consider uncertainty in the learning predictions, they lack natural mechanisms to handle sensor noise or other sources of uncertainty. In~\Cref{SS:active_learning}, we discuss how active learning (AL) techniques can mitigate some of these issues.
\begin{table*}
  \centering
  \begin{minipage}{.99\textwidth}
  {\footnotesize
  \begin{tabular}{@{}c p{0.53\textwidth} p{0.4\textwidth}@{}}
    \toprule
     & \multicolumn{1}{l}{Algorithm keyword} & \multicolumn{1}{l}{References} \\ \midrule
     \multirow{9}{*}{\STAB{\rotatebox[origin=c]{90}{Architecture/method}}}
     & Conditional variational autoencoder      & \citep{schmid2022fast}     \\
     & Convolutional neural network  & \citep{schmid2022fast,bai2017toward,ly2019autonomous,caley2020environment,caley2019deep,zwecher2022integrating, hepp2018learn,vasquez2021next,mendoza2020supervised,li2023learning,saroya2020online,georgakis2022upen,ramakrishnan2020occupancy,li2023learning,pan2023views,pan2023oneshot} \\
     & Gaussian process     &  \citep{rayas2022informative,cao2023catnipp,ruckin2022adaptive,yang2023intent,wang2023spatiotemporal,marchant2014sequential,hitz2017adaptive,popovic2020informative,hollinger2014sampling,vivaldini2019uav,viseras2019deepig,yanes2023deep,choi2021adaptive,wei2020informative,song2015trajectory,denniston2023learned, ott2024approximate, ott2024trajectory} \\
     & Implicit neural representation     &  \citep{jin2023neunbv,pan2022activenerf,ran2022neurar,zhan2022activermap,sunderhauf2023density,yan2023active,lee2022uncertainty,zhang2023affordance,pan2023views} \\
     & Kernel density estimation      &  \citep{zacchini2023informed}   \\
     & Occupancy prediction network       & \citep{tao2023seer} \\
     & Scene completion network       & \citep{schmid2022scexplorer} \\
     & Transformer       & \citep{dhami2023pred,hanlon2023active} \\
     & Variational autoencoder      &  \citep{shrestha2019learned}    \\
     \midrule
     
     \multirow{14}{*}{\STAB{\rotatebox[origin=c]{90}{Training objective}}}
     & 3D semantic scene completion       & \citep{schmid2022scexplorer}       \\
     & 2D occupancy map completion       & \citep{shrestha2019learned,zwecher2022integrating,georgakis2022upen,ramakrishnan2020occupancy}     \\
     & 3D occupancy map completion       & \citep{tao2023seer}     \\
     & Environment topology      &  \citep{saroya2020online}     \\
     & Environmental variable      &  \citep{rayas2022informative,cao2023catnipp,ruckin2022adaptive,yang2023intent,wang2023spatiotemporal,marchant2014sequential,hitz2017adaptive,popovic2020informative,hollinger2014sampling,vivaldini2019uav,popovic2020informativelocunc,viseras2019deepig,yanes2023deep,choi2021adaptive,caley2020environment,wei2020informative,song2015trajectory,denniston2023learned, ott2024approximate, ott2024trajectory} \\
     & Grasp affordance      &  \citep{zhang2023affordance}     \\
     & Informative view distribution      &  \citep{schmid2022fast,zacchini2023informed}     \\
     & Localization score      &  \citep{hanlon2023active}     \\
     & Minimum view subset for coverage      &  \citep{pan2023oneshot}  \\
     & Next-best view      &  \citep{vasquez2021next,mendoza2020supervised}  \\
     & Point cloud shape completion      &  \citep{dhami2023pred}  \\
     & Radiance field      &  \citep{jin2023neunbv,pan2022activenerf,zhang2023affordance,ran2022neurar,zhan2022activermap,lee2022uncertainty,sunderhauf2023density,yan2023active}     \\
     & Required number of views      &  \citep{pan2023views}  \\
     & Target point of interest      &  \citep{caley2019deep} \\
     & Unexplored navigable area beyond map frontiers      &  \citep{li2023learning} \\
     & Utility/reward      &  \citep{hepp2018learn,schmid2022fast,bai2017toward,ly2019autonomous}  \\ \midrule
     
     \multirow{16}{*}{\STAB{\rotatebox[origin=c]{90}{Mission objective}}}
     & Coverage     & \citep{schmid2022scexplorer,schmid2022fast,saroya2020online,ly2019autonomous,vasquez2021next,mendoza2020supervised,zwecher2022integrating,tao2023seer,li2023learning}    \\
     & Estimate quantile values         &  \citep{rayas2022informative}    \\
     & Find maximum field value         &  \citep{song2015trajectory,denniston2023learned}    \\
     & Maximize grasp quality & \citep{zhang2023affordance} \\
     & Maximize localization accuracy & \citep{hanlon2023active} \\
     & Maximize probability of improvement/expected improvement         &  \citep{denniston2023learned, ott2024approximate, ott2024trajectory}    \\
     & Maximize observed surfaces      &  \citep{dhami2023pred,pan2023oneshot}       \\
     & Minimize explored area to target point of interest & \citep{caley2019deep} \\
     & Minimize map error      & \citep{viseras2019deepig,choi2021adaptive,caley2020environment,ramakrishnan2020occupancy}       \\
     & Minimize map uncertainty                       &  \citep{hepp2018learn,bai2017toward,shrestha2019learned,zacchini2023informed,hollinger2014sampling,yanes2023deep,georgakis2022upen,wei2020informative,denniston2023learned, ott2024approximate, ott2024trajectory} \\
     & Minimize map uncertainty and robot localization uncertainty      & \citep{popovic2020informativelocunc}  \\
     & Minimize map uncertainty in high-interest areas           & \citep{cao2023catnipp,ruckin2022adaptive,yang2023intent,marchant2014sequential,hitz2017adaptive,popovic2020informative,hollinger2014sampling}    \\
     & Minimize map uncertainty in unclassified areas      & \citep{vivaldini2019uav}       \\
     & Minimize model uncertainty & \citep{jin2023neunbv,pan2022activenerf,ran2022neurar,zhan2022activermap,lee2022uncertainty,sunderhauf2023density,yan2023active} \\ 
     & Minimize target localization uncertainty      &  \citep{wang2023spatiotemporal}       \\
     & Object reconstruction based on PSNR     &  \citep{pan2023views}       \\
     & Point-goal navigation & \citep{georgakis2022upen,ramakrishnan2020occupancy} \\  \midrule
     
     \multirow{17}{*}{\STAB{\rotatebox[origin=c]{90}{Training data}}}
     & 3D Street View dataset      &  \citep{hepp2018learn}   \\
     & Custom CAD models                      &  \citep{vasquez2021next,mendoza2020supervised,ran2022neurar,yan2023active,pan2023oneshot} \\
     & Custom synthetic dataset         &  \citep{schmid2022fast,bai2017toward,saroya2020online,cao2023catnipp,ruckin2022adaptive,yang2023intent,hitz2017adaptive,popovic2020informative,popovic2020informativelocunc,viseras2019deepig,zwecher2022integrating,song2015trajectory,marchant2014sequential}   \\
     & DTU dataset      &  \citep{jin2023neunbv}   \\
     & Heuristic parameter setting                     &  \citep{wang2023spatiotemporal,yanes2023deep} \\
     & Gibson dataset      &  \citep{ramakrishnan2020occupancy}   \\
     & INRIA Aerial Image Labeling dataset                & \citep{ly2019autonomous}     \\
     & KTH dataset                 & \citep{shrestha2019learned,caley2019deep}    \\
     & LLFF dataset      &  \citep{pan2022activenerf,sunderhauf2023density}   \\
     & Matterport3D dataset      &  \citep{georgakis2022upen,tao2023seer,li2023learning,hanlon2023active}   \\
     & NeRF/Synthetic-NeRF dataset      &  \citep{pan2022activenerf,ran2022neurar,lee2022uncertainty,sunderhauf2023density,zhan2022activermap}   \\
     & NYU dataset       & \citep{schmid2022scexplorer}     \\
     & Real-world environmental data                      &  \citep{rayas2022informative,hitz2017adaptive,hollinger2014sampling,caley2019deep,wei2020informative,denniston2023learned,vivaldini2019uav} \\ 
     & Realistic simulator                      &  \citep{zacchini2023informed,zhang2023affordance} \\
     & Regional Ocean Modeling System (ROMS) dataset                & \citep{caley2020environment}     \\   
     & ShapeNet dataset      &  \citep{dhami2023pred,jin2023neunbv,pan2023views}   \\
     & Tanks\&Temples dataset      &  \citep{zhan2022activermap}   \\ \bottomrule
  \end{tabular}}
  \end{minipage} 
 \caption{Supervised learning methods for \ac{AIPP}.\label{T:supervised_learning}} 
\end{table*}


\subsection{Reinforcement Learning} \label{SS:deep_rl}
\Ac{RL} is a branch of machine learning that equips an agent to learn from an environment by interacting with it and receiving feedback in terms of rewards or penalties. In \ac{AIPP}, \ac{RL} has emerged as an effective learning methodology, enabling a system to adapt its behavior based on an evolving understanding of its environment.

The problem of \ac{AIPP} can be framed as a \ac{POMDP} ~\citep{kochenderfer2022algorithms}. Formally, a \ac{POMDP} is defined by the tuple $\langle \mathcal{S}, \mathcal{A}, \mathcal{O}, T, Z, R \rangle$, where:

\begin{itemize}
    \item $\mathcal{S}$ is the \textit{state} space;
    \item $\mathcal{A}$ is the \textit{action} space;
    \item $\mathcal{O}$ is the \textit{observation} space;
    \item $T(s' \mid s,a)$ is the \textit{state transition function};
    \item $Z(o \mid s',a)$ is the \textit{observation function};
    \item $R$ is the immediate \textit{reward function}.
\end{itemize}

In the context of \ac{AIPP}, the true state of the environment is the actual physical attributes of the environment captured by $F(\mathbf{x})$ in~\Cref{S:formulation}, e.g., the locations of obstacles and areas of interest, while the observation $z$ is the robot's current measurement of these attributes based on sensor readings.

The \textit{state} space $\mathcal{S}$ is defined as the set of all possible feature mappings $F$ in the environment and agent-specific information, e.g., its position or battery level. The \textit{action} space $\mathcal{A}$ consists of all possible actions $a_i \in \mathcal{A}$ that the robot could take, e.g., moving to a given location, using a specific sensor, or returning to a charging station. The \textit{observation} space $\mathcal{O}$ corresponds to the set of all possible sensor readings and derived interpretations the robot could have about the environment, e.g., semantic segmentation of RGB images.

The \textit{state transition function} $T(s' \,|\, s, a)$ describes the probabilistic dynamics of the robot's state in the context of the \ac{AIPP} problem. When the robot executes an action $a \in \mathcal{A}$ in its current state $s \in \mathcal{S}$, it moves to a new state $s' \in \mathcal{S}$. The inherent uncertainty of action outcomes and their impact on the environment is captured by the transition function. This function is used in the belief update, which refines the robot's understanding of the environment feature mapping $F$. Additionally, the \textit{observation function} $Z(o \,|\, s', a)$ provides the likelihood of perceiving an observation $o \in \mathcal{O}$ subsequent to action $a$ and arriving at state $s'$. This observation model, which is probabilistically linked to the feature mapping $F$ through $z_t \sim p(z \,|\, F)$, is used to update the robot's belief state using~\Cref{eq:map_belief}. 


The \textit{reward function} $R(s,a)$ captures the trade-off between maximizing information gain, as described by the criterion $I(\cdot)$, and adhering to a budget constraint $C(\psi) \leq B$. While the objective function $I(\cdot)$ evaluates a sequence of actions, the \textit{reward function} $R(s,a)$ evaluates individual actions. That is, the information criterion $I(\cdot)$ in \ac{AIPP} accumulates the rewards $R(s,a)$ over a sequence of actions, and can additionally incorporate a discount factor $\gamma$ to reflect that future rewards are less valuable than immediate rewards: \begin{equation} \label{eq:rl_return}
    I(\psi) = \sum_{t=1}^{N} \gamma^{t-1} R(s_t,a_t) \, .
\end{equation} In \ac{AIPP}, the discount factor is typically set to $\gamma = 1$ for finite horizons with budget constraints. 

The \textit{policy} is a function $\pi : \mathcal{S} \rightarrow \mathcal{A}$. The concept of a policy and the general action sequence $\psi$ in \Cref{eq:ipp_problem} are linked by $\psi = (\pi(s_1), \dots, \pi(s_N))$, where the states evolve based on the previous state and action according to $T(s' \,|\, s, a)$. The optimal policy can be queried to select an action sequence $\psi^*$ maximizing the total expected discounted return, and hence the information gain, while being within the allowed budget. 


Any \ac{POMDP} can be viewed as an MDP that uses beliefs as states, also called a belief-state MDP or belief MDP. The \textit{state} space of a belief MDP is the set of all beliefs $\mathcal{B}$. The \textit{action} space is equivalent to that of the POMDP~\citep{kochenderfer2022algorithms}. To connect this directly to~\Cref{S:background}, in \ac{AIPP}, the robot’s belief about the true state of the environment is denoted by $\tilde{F}$ in~\Cref{SS:mapping}. This belief evolves as the robot gathers information about its environment, which starts as partially unknown and is gradually discovered through sensing and exploration. 

\subsubsection{Network Architectures and Training Algorithms}
\edit{Various neural network architectures can be used in \ac{RL} for \ac{AIPP}, such as attention-based neural networks, convolutional neural networks (CNN), long short-term memory (LSTM) networks, and graph neural networks (GNN).}

\edit{Attention-based neural networks, inspired by human visual attention, focus selectively on input data, which is useful for \ac{AIPP} where the robot focuses on high-interest regions. CNNs excel at capturing spatial hierarchies in 2D occupancy grid maps. LSTMs, a type of recurrent neural network, learn long-term dependencies, which is ideal for temporally dependent problems in AIPP. 
GNNs operate on graph structures, making them most appropriate for graph-modeled environments.} 

\edit{Several \ac{RL} algorithms can train these network architectures including proximal policy optimization (PPO), asynchronous actor-critic (A3C), soft actor-critic (SAC), advantage actor-critic (A2C), deep Q-network (DQN), double deep Q-network (DDQN), as well as many other variations and combinations, e.g. AlphaZero. These algorithms vary in exploration and exploitation, sample efficiency, computational needs, and stability.}

\edit{Actor-critic methods use a value function estimate to guide optimization. The actor is the policy, while the critic is the value function. They train in parallel, differing in value function, advantage function, or action-value function approximation. Most methods focus on stochastic policies, but some support deterministic continuous actions. For example, PPO, A3C, SAC, and A2C follow an actor-critic approach with separate policy and value networks. In contrast, DQN and DDQN are critic-only value-based methods, where the optimal policy is inferred from the value function.}

\edit{AlphaZero combines the actor-critic framework with Monte Carlo tree search (MCTS), using the network's policy to guide the search and the value function to evaluate leaf nodes. MCTS simulations generate a robust policy, considering a longer planning horizon \citep{silver2017mastering, silver2018general}.}

\edit{Various open-source implementations and benchmarks of these network architectures and training algorithms are widely available ~\citep{stable-baselines, deepmind2020jax, egorov2017pomdps, Ansel_PyTorch_2_Faster_2024, Abadi_TensorFlow_Large-scale_machine_2015}.}

\subsubsection{Reinforcement Learning Approaches}
A variety of \ac{RL}-based approaches have been explored in the literature. In~\Cref{T:reinforcement_learning}, we distinguish the methods primarily by how they formulate each component of the POMDP architecture as well as how they learn policies. 

All of the methods surveyed include some aspect of the current robot information in the state. The robot pose is often used~\citep{cao2023adriadne, lodel2022where, zeng2022deep, marchant2014sequential, denniston2023learned}, while other methods include additional robot information, e.g., energy level~\citep{choudhury2020adaptive, ott2023sboaippms}, viewing direction~\citep{chen2022learning}, and operational status~\citep{bayerlein2021multi}. In addition to current robot information, all methods store some form of a spatial map in the state. The most popular spatial maps are Gaussian processes~\citep{cao2023catnipp, wang2023spatiotemporal, yang2023intent, wei2020informative, ruckin2022adaptive, ott2023sboaippms, yanes2023deep, viseras2019deepig, choi2021adaptive, marchant2014sequential, denniston2023learned} and occupancy grids~\citep{yang2023learning, niroui2019deep, chen2020autonomous, lodel2022where, chen2019self, westheider2023multi, liu2023learning, gao2022cooperative, zwecher2022integrating, chen2022learning, chen2021zeroshot, cao2023adriadne, sukkar2019multi}, both of which are described in~\Cref{SS:mapping}. Another common component of the state is the previous position and sensor history. The history components can range from the executed trajectory~\citep{cao2023catnipp} to the previous graph history~\citep{wang2023spatiotemporal} as well as the previous observations~\citep{viseras2019deepig, denniston2023fast}. Some methods implicitly capture the history of robot poses and measurements in the belief state action-observation history~\citep{choudhury2020adaptive, ott2023sboaippms} or by capturing the previously explored areas in the occupancy grid~\citep{chen2021zeroshot, westheider2023multi, gao2022cooperative}. 

The \textit{action} space determines the set of actions available to the robot in the current state. At the highest level, actions are distinguished by being discrete (e.g., move left or move right) or continuous (e.g., set velocity to \SI{5}{\meter/\second}). For example, \citet{yang2023learning} use a continuous action space consisting of longitudinal and lateral velocity commands while \citet{lodel2022where} use a continuous 2D reference viewpoint for their local planner. \citet{huttenrauch2019deep} use linear and angular velocity or acceleration commands. Continuous action spaces are not just restricted to movement commands. \citet{bartolomei2021semantic} focus on semantic-aware active perception while reaching a goal target using an action space consisting of a set of weights to assign to each semantic class. 

Continuous action spaces often significantly increase the computational complexity of \ac{POMDP}s. In many cases, discrete action spaces are simpler to work with. When the spatial map is represented as an abstract graph, a common choice is for movement commands to be represented as movement from one graph node to another via the connecting edge~\citep{cao2023catnipp, cao2023adriadne, wang2023spatiotemporal, yang2023intent, wei2020informative, ruckin2022adaptive, ott2023sboaippms, choudhury2020adaptive, westheider2023multi, yanes2023deep, viseras2019deepig, viseras2021wildfire, bayerlein2021multi, denniston2023fast, duecker2021embedded, denniston2023learned, gao2022cooperative, best2019dec, zwecher2022integrating}. Another common approach relies on frontiers, which represent the boundary between known and unknown space in the environment and can be seen as a special case of a graph node action. The use of frontiers as movement actions does not necessarily require the spatial map to be represented as an abstract graph~\citep{niroui2019deep, chen2020autonomous, chen2021zeroshot, liu2023learning}. Other discrete actions include selecting from a discrete set of sensors to employ~\citep{choudhury2020adaptive, ott2023sboaippms} as well as motion primitive commands~\citep{marchant2014sequential, song2015trajectory} which can be seen as a discretization of continuous actions where the agent is given a set of potential trajectories to execute which correspond to actions. 

The \textit{reward function} is the most problem-dependent component of the formulation. One of the core challenges that arise in the \ac{POMDP} framework and, consequently, in \ac{AIPP}, is the exploration-exploitation trade-off. The agent must select actions that balance between exploring unknown areas of the environment to increase its knowledge (exploration) and exploiting its current knowledge to find areas of interest within the constraints of its available resources (exploitation). This trade-off is intrinsically linked to the \ac{AIPP} objective of maximizing the information criterion while respecting the cost constraint. Hence, it is a critical factor driving the learning process of the \ac{RL} agent. Essentially, all \ac{AIPP} methods use some measure of informativeness as the objective as discussed in~\Cref{SS:evaluation}. Map uncertainty reduction in high-interest areas is a very common \ac{RL} reward in the \ac{AIPP} literature~\citep{yang2023intent, ott2023sboaippms, ruckin2022adaptive, westheider2023multi, viseras2021wildfire, denniston2023fast, marchant2014sequential, duecker2021embedded, denniston2023learned, song2015trajectory}. 
Some approaches use only map uncertainty reduction to spread measurements across the environment as uniformly as possible and minimize the estimation error~\citep{cao2023catnipp, niroui2019deep, wang2023spatiotemporal, ott2023sboaippms, yanes2023deep, viseras2019deepig, choi2021adaptive, cao2023adriadne}. Coverage is a common surrogate metric for map uncertainty reduction since it only relies on computing the area uncovered as the robot travels through the environment, which is often faster than computing the Gaussian process variance online~\citep{zeng2022deep, liu2023learning, kumar2022graph, bayerlein2021multi, sukkar2019multi, best2019dec, zwecher2022integrating, chen2022learning}. While coverage can perform well on the map uncertainty reduction objective, it cannot explicitly reason about high-interest areas. The mutual information between observations and landmarks is another common way to measure map uncertainty reduction~\citep{yang2023learning, lodel2022where, wei2020informative, chen2019self}. Some methods also explicitly incorporate distance-based metrics in the reward function~\citep{chen2020autonomous, chen2021zeroshot, cao2023adriadne, lodel2022where, gao2022cooperative}. For example, \citet{chen2020autonomous, chen2021zeroshot} and \citet{cao2023adriadne} assign greater rewards for actions that achieve the same uncertainty reduction in less distance.

Optimally solving \ac{POMDP}s is computationally challenging~\citep{kaelbling1998planning, kochenderfer2022algorithms}. However, practical solutions can be obtained through various \ac{RL} algorithms. We categorize \ac{AIPP} policies based on their deployment inference strategies, distinguishing between zero-shot inference and expected discounted $m$-step search policies. Zero-shot inference policies are trained offline, considering the sequence of actions. The policy is trained to account for the cascading impact of each action on subsequent decisions, taking into consideration the consequences of actions and future dependencies during the offline training process. Upon deployment, these policies take the current belief state---updated to reflect the cumulative knowledge of the environment as per~\Cref{eq:map_belief}---and directly output the next best action without additional planning. Conversely, policies using an expected discounted $m$-step search are dynamic in the sense that they actively consider a sequence of potential future actions during deployment. Instead of reacting to the current belief state, this approach projects forward, evaluating a decision tree of potential action sequences up to $m$ steps ahead and factoring in the anticipated updates to the belief state for each action. The policy then chooses an action based on the strategy that maximizes the total expected reward over this lookahead horizon.


\begin{table*}[!h]
  \centering
  \begin{minipage}{.99\textwidth}
  {\renewcommand{\arraystretch}{1.15}
  \begin{tabular}{@{}c p{0.4\textwidth} p{0.53\textwidth}@{}}
    \toprule
     & \multicolumn{1}{c}{Description} & \multicolumn{1}{c}{References} \\ \midrule
     \multirow{6}{*}{\STAB{\rotatebox[origin=c]{90}{State}}}
     & Current robot information  & \cite{yang2023learning, cao2023catnipp, niroui2019deep, chen2020autonomous, chen2021zeroshot, cao2023adriadne, lodel2022where, wang2023spatiotemporal, yang2023intent, zeng2022deep, wei2020informative, ruckin2022adaptive, chen2019self, ott2023sboaippms, choudhury2020adaptive, westheider2023multi, yanes2023deep, liu2023learning, viseras2019deepig, viseras2021wildfire, bartolomei2021semantic, kumar2022graph, choi2021adaptive, bayerlein2021multi, gazani2023bag, denniston2023fast, marchant2014sequential, duecker2021embedded, denniston2023learned, gao2022cooperative, sukkar2019multi, best2019dec, huttenrauch2019deep, zwecher2022integrating, chen2022learning, qie2019joint, fan2020distributed, arora2019multi,bucher2021adversiarial, singh2009nonmyopic, ott2022risk, bouman2022adaptive} \\
     & Observation history & \cite{ye2018active, lodel2022where, wang2023spatiotemporal, choudhury2020adaptive, westheider2023multi, liu2023learning, viseras2019deepig, gazani2023bag, denniston2023fast,bucher2021adversiarial, singh2009nonmyopic} \\
     & Robot history  & \cite{cao2023catnipp, chen2020autonomous, chen2021zeroshot, wang2023spatiotemporal, gazani2023bag, denniston2023fast,  gao2022cooperative} \\
     & Spatial map  & \cite{yang2023learning, cao2023catnipp, niroui2019deep, chen2020autonomous, chen2021zeroshot, cao2023adriadne, lodel2022where, wang2023spatiotemporal, yang2023intent, zeng2022deep, wei2020informative, ruckin2022adaptive, chen2019self, ott2023sboaippms, choudhury2020adaptive, westheider2023multi, yanes2023deep, liu2023learning, viseras2019deepig, viseras2021wildfire, bartolomei2021semantic, kumar2022graph, choi2021adaptive, bayerlein2021multi, gazani2023bag, denniston2023fast, marchant2014sequential, duecker2021embedded, denniston2023learned, gao2022cooperative, sukkar2019multi, best2019dec, huttenrauch2019deep, zwecher2022integrating, chen2022learning, qie2019joint, fan2020distributed, arora2019multi, singh2009nonmyopic, ott2022risk, bouman2022adaptive} \\ \midrule
     \multirow{9}{*}{\STAB{\rotatebox[origin=c]{90}{Map}}}
     & 2D Gaussian process & \cite{cao2023catnipp, wang2023spatiotemporal, yang2023intent, wei2020informative, ruckin2022adaptive, ott2023sboaippms, yanes2023deep, viseras2019deepig, choi2021adaptive, marchant2014sequential, denniston2023learned} \\
     & 2D occupancy grid &  \cite{yang2023learning, niroui2019deep, chen2020autonomous, lodel2022where, chen2019self, westheider2023multi, liu2023learning, gao2022cooperative, zwecher2022integrating, chen2022learning, ott2022risk, bouman2022adaptive} \\
     & 3D occupancy grid & \cite{chen2021zeroshot, cao2023adriadne, sukkar2019multi} \\
     & 3D semantic occupancy map & \citep{zeng2022deep, bartolomei2021semantic} \\
     & Bayesian network & \cite{arora2019multi} \\
     & Factor graph & \cite{denniston2023fast}      \\
     & Gaussian Markov random field & \cite{duecker2021embedded} \\
     & Landmark locations and uncertainties   &   \cite{chen2020autonomous, chen2021zeroshot, bartolomei2021semantic, fan2020distributed}                        \\
     & Location graph  & \cite{best2019dec, choudhury2020adaptive, huttenrauch2019deep, qie2019joint, singh2009nonmyopic, ott2022risk, bouman2022adaptive} \\
     \midrule
     \multirow{4}{*}{\STAB{\rotatebox[origin=c]{90}{Action}}}
     & Continuous action & \cite{yang2023learning, lodel2022where, bartolomei2021semantic, sukkar2019multi, huttenrauch2019deep, fan2020distributed, denniston2023learned,bucher2021adversiarial} \\
     & Discrete action & \cite{zeng2022deep, chen2019self, ott2023sboaippms, viseras2021wildfire, kumar2022graph, choi2021adaptive, bayerlein2021multi, gazani2023bag, marchant2014sequential, chen2022learning, qie2019joint, arora2019multi, ye2018active, ott2022risk} \\
     & Frontier & \cite{niroui2019deep, chen2020autonomous, chen2021zeroshot, liu2023learning} \\
     & Graph node & \cite{cao2023catnipp, cao2023adriadne, wang2023spatiotemporal, yang2023intent, wei2020informative, ruckin2022adaptive, ott2023sboaippms, choudhury2020adaptive, westheider2023multi, yanes2023deep, viseras2019deepig, viseras2021wildfire, bayerlein2021multi, denniston2023fast, duecker2021embedded, denniston2023learned, gao2022cooperative, best2019dec, zwecher2022integrating, singh2009nonmyopic, bouman2022adaptive} \\ \midrule
     \multirow{3}{*}{\STAB{\rotatebox[origin=c]{90}{Policy}}}
     & Expected discounted $m$-step reward & \cite{ruckin2022adaptive, ott2023sboaippms, choudhury2020adaptive, denniston2023fast, marchant2014sequential, duecker2021embedded, denniston2023learned, sukkar2019multi, best2019dec, arora2019multi, bucher2021adversiarial, ott2022risk, bouman2022adaptive} \\
     & Zero-shot inference & \cite{yang2023learning, cao2023catnipp, niroui2019deep, chen2020autonomous, chen2021zeroshot, cao2023adriadne, lodel2022where, wang2023spatiotemporal, yang2023intent, zeng2022deep, wei2020informative, chen2019self, westheider2023multi, yanes2023deep, liu2023learning, viseras2019deepig, viseras2021wildfire, bartolomei2021semantic, kumar2022graph, choi2021adaptive, bayerlein2021multi, gazani2023bag, gao2022cooperative, huttenrauch2019deep, zwecher2022integrating, chen2022learning, qie2019joint, fan2020distributed, denniston2023learned, ye2018active, singh2009nonmyopic} \\
     \midrule 
     \multirow{11}{*}{\STAB{\rotatebox[origin=c]{90}{Reward}}}
     & Belief state mode & \cite{choudhury2020adaptive} \\
     & Coverage & \cite{zeng2022deep, liu2023learning, kumar2022graph, bayerlein2021multi, sukkar2019multi, best2019dec, zwecher2022integrating, chen2022learning, bucher2021adversiarial, ott2022risk, bouman2022adaptive} \\
     & Maximize inter-agent distance & \citep{huttenrauch2019deep} \\
     & Maximize mutual information of observations and landmarks  & \cite{yang2023learning, lodel2022where, wei2020informative, chen2019self} \\
     & Minimize landmark localization uncertainty & \cite{chen2020autonomous, chen2021zeroshot, bartolomei2021semantic, gazani2023bag} \\
     & Minimize map uncertainty & \cite{cao2023catnipp, niroui2019deep, wang2023spatiotemporal, ott2023sboaippms, yanes2023deep, viseras2019deepig, choi2021adaptive, cao2023adriadne} \\
     & Minimize map uncertainty in high-interest areas & \cite{yang2023intent, ott2023sboaippms, ruckin2022adaptive, westheider2023multi, viseras2021wildfire, denniston2023fast, marchant2014sequential, duecker2021embedded, denniston2023learned, arora2019multi, singh2009nonmyopic}  \\ 
     & Size of bounding box for target detection & \cite{ye2018active} \\
     & Travel distance & \cite{chen2020autonomous, chen2021zeroshot, cao2023adriadne, lodel2022where, gao2022cooperative, qie2019joint}\\
     & Travel time & \cite{fan2020distributed} \\
     \bottomrule     
  \end{tabular}}
  \caption{Reinforcement learning methods for \ac{AIPP}.\label{T:reinforcement_learning}}
  \end{minipage}
\end{table*}

\citet{yang2023learning} use an attention-based \ac{MLP} which is trained using \ac{PPO} to produce a zero-shot inference action of continuous longitudinal and lateral velocity with the goal of maximizing the mutual information between observations and landmark states. \citet{cao2023catnipp} take a slightly different approach by using an attention-based graph neural network and \ac{LSTM} network with \ac{PPO} to select the next graph node. \citet{ruckin2022adaptive} use AlphaZero by combining Monte Carlo tree search (MCTS) with a \ac{CNN} to learn information-rich actions in adaptive data gathering missions. \citet{chen2019self} uses deep Q-network (DQN) with a \ac{CNN} to select the next location to visit from a set of randomly sampled 2D positions with the goal of maximizing the mutual information between received observations and the occupancy grid map. \citet{chen2021zeroshot} uses the advantage actor-critic (A2C) algorithm with a graph neural network to select the next best frontier to visit with the goal of maximizing landmark uncertainty reduction and minimizing travel distance. \citet{cao2023adriadne} uses the soft actor-critic (SAC) algorithm with an attention-based graph neural network to output the next graph node to visit with the goal of maximizing the number of observed frontiers while minimizing travel distance. Other methods mainly differ in the training algorithms and the network architectures used with the most popular being PPO, A2C, SAC, and CNN, DQN, LSTM, respectively~\citep{yang2023learning, cao2023catnipp, niroui2019deep, chen2020autonomous, chen2021zeroshot, cao2023adriadne, lodel2022where, wang2023spatiotemporal, chen2019self, liu2023learning, gazani2023bag, zwecher2022integrating}.

In summary, \ac{RL} offers a powerful framework for \ac{AIPP}. By learning to optimize actions for maximum information gain, \ac{RL}-based systems can adaptively and effectively explore unknown environments. However, \ac{RL} algorithms also face challenges such as data-intensive training, sensitivity to reward design, and computational demands, which can affect their efficiency and generalizability in real-world robotics scenarios. Existing \ac{RL}-based \ac{AIPP} literature addresses these issues through advanced neural network architectures, sophisticated training algorithms, careful reward and policy design, and robust map representations. Additionally, most of these approaches are tested on environments very similar to those used for training. A promising direction for future work is thus to develop methods that can generalize to new environments.

\subsection{Imitation Learning} \label{SS:imitation_learning}
Reinforcement learning (RL) approaches typically assume that either the \textit{reward function} is known or that rewards are received while interacting with the environment. For some applications, it may be easier for an expert to demonstrate the desired behavior rather than specify a reward function. \Ac{IL} refers to the case where the desired behavior is learned from expert demonstrations. 

A simple form of \ac{IL} can be thought of as a supervised learning problem from~\Cref{SS:supervised_learning} typically referred to as behavioral cloning \cite{pomerleau1991efficient}. The objective is to train a stochastic policy $\pi: \mathcal{S} \to \mathcal{A}$ where the policy $\pi_{\mathbf{\theta}}$ is parameterized by $\mathbf{\theta}$. The training objective is to maximize the likelihood of actions from a dataset $\mathcal{D}$ of expert state-action pairs: \begin{equation}
  \underset{\theta}{\mathrm{maximize}} \prod_{(s,a) \in D} \pi_{\theta}(a \mid s) . \label{eq:behavior_cloning}
\end{equation}

This approach, while straightforward, is limited by the quality and quantity of the expert demonstrations available. Moreover, it often generalizes poorly to states not encountered in the training dataset, leading to suboptimal or even erroneous behaviors under novel circumstances. More sophisticated \ac{IL} strategies have been developed to address these limitations. These include methods combining \ac{IL} with \ac{RL} to refine policies beyond the initial demonstrations, approaches that incorporate active learning (AL) to query experts for demonstrations in states where the learner is uncertain, and algorithms that utilize inverse \ac{RL} to infer the underlying reward function implied by the expert's behavior.

It is important to note the similarities and differences between \ac{IL} and \ac{RL}. Both \ac{IL} and \ac{RL} seek to find an optimal policy for a given task. However, the key distinction lies in how they learn this policy. \ac{RL} aims to learn the policy by maximizing the expected discounted return, as illustrated in~\Cref{eq:rl_return}. In contrast, \ac{IL}, as depicted in~\Cref{eq:behavior_cloning}, assumes access to samples from an expert policy and focuses on mimicking this behavior. This approach treats the expert policy as a black box, learning to replicate its decisions without necessarily understanding the underlying motivations. In \ac{IL}, the reliance on expert demonstrations aligns closely with the concept of an \ac{AL} oracle discussed in~\Cref{SS:active_learning}, which provides guidance in the form of expert input, albeit in a more passive manner.

\ac{IL} has been effectively applied in robotic scenarios, including indoor autonomous exploration, path planning, and multi-robot coordination. \citet{liu2023learning} introduce the Learning to Explore (L2E) framework, combining IL with deep \ac{RL} for \acp{UGV} in indoor environments. This approach pretrains exploration policies using \ac{IL} and then refines them with \ac{RL}, leading to improved sample efficiency and training speed and outperforming heuristic methods in diverse environments.

\citet{choudhury2018data} propose a data-driven \ac{IL} framework to train planning policies by imitating a clairvoyant oracle with ground truth knowledge of the world map. This approach efficiently trains policies based on partial information for tasks like \ac{AIPP} and motion planning, showing significant performance gains over state-of-the-art algorithms, even in real \ac{UAV} applications. \citet{reinhart2020learning} develop a learning-based path planning approach for \acp{UAV} in unknown subterranean environments. They use a graph-based path planner as a training expert for \ac{IL}, resulting in a policy that guides autonomous exploration with reduced computational costs and less reliance on consistent, online reconstructed maps.

\citet{li2020graph} address decentralized multi-robot path planning by proposing a combined model that synthesizes local communication and decision-making policies. Their architecture, comprising a convolutional neural network and a graph neural network, was trained to imitate an expert. This model demonstrates its effectiveness in decentralized planning with local communication and observations in cluttered workspaces. \citet{tzes2023graph} tackles multi-robot \ac{AIPP} with their Information-aware Graph Block Network (I-GBNet). Trained via IL, this network aggregates information over a communication graph and offers sequential decision-making in a distributed manner. Its scalability and robustness were validated in large-scale experiments involving dynamic target tracking and localization. Lastly, \citet{dai2022camera} explores indoor environments using an RGB-D camera and generative adversarial imitation learning (GAIL). They modified the ORB-SLAM2 method for enhanced navigation and trained a camera view planning policy with GAIL, yielding efficient exploration and reducing tracking failure in indoor environments.

These studies show the versatility and effectiveness of \ac{IL} in a wide range of robotic applications, demonstrating its potential to enhance efficiency, decision-making, and adaptability in varied and challenging environments.  However, significant challenges and areas for future research remain. Key challenges include improving the generalizability of learned behaviors to unseen environments, enhancing the robustness of \ac{IL} algorithms against dynamic and unpredictable real-world conditions, and developing more efficient methods for collecting and using expert demonstrations. Further work is also needed to integrate \ac{IL} with other learning paradigms, such as \ac{RL}, to address complex multi-agent systems and high-dimensional state spaces. 


\subsection{Active Learning} \label{SS:active_learning}
The problem of \ac{AIPP} can also be seen as an active learning (AL) process, whereby the learner uses information measures to choose actions for collecting useful or descriptive data~\citep{taylor2021active}. Here, we mathematically connect the general concept of \ac{AL} to robotic \ac{AIPP} theory. We then draw parallels between different components of \ac{AL} and the learning-based \ac{AIPP} methods presented in the previous subsections. Lastly, we elaborate on an emerging line of work in \ac{AIPP} that uses \ac{AL} objectives to improve sensor models in new environments.

\ac{AL} addresses the task of finding the most informative data in a set of unlabeled samples such that a model is maximally improved if that sample is labeled and added to the existing training dataset~\citep{settles2009active}. We define the \ac{AL} problem as follows:
\begin{itemize}
    \item $\mathcal{Q}$ is the space of \textit{queries} $q_i \in \mathcal{Q}$ and $N_q \in \mathbb{N}$ is the number of sampled queries;
    \item $\mathcal{L}$ is the space of \textit{targets} $l_i \in \mathcal{L}$ corresponding to query $q_i$;
    \item $\theta: \mathcal{Q} \to \mathcal{L}$ is the \textit{model} predicting a target $\theta(q_i) = l_i$ given a query $q_i$;
    \item $P: \Theta \to \mathbb{R}$ is the \textit{performance function} evaluating a model $\theta \in \Theta$;
    \item $O: \mathcal{Q} \to \mathcal{L}$ is an \textit{oracle} that can be queried to generate a target $O(q_i) = l_i$ for a specific query $q_i$;
    \item $q = (q_1, \ldots, q_i) \in \mathcal{Q}^i$, $l = (O(q_1), \ldots, O(q_i)) \in \mathcal{L}^i$ are collected queries and targets up to iteration $i \leq N_q$;
    \item $A: \Theta \times \mathcal{Q} \to \mathbb{R}$ is the \textit{acquisition function} estimating a query's $q_i$ effect on the model performance $P(\theta)$ after re-training on collected training data pairs $q$ and $l$.
\end{itemize}

We aim to maximize the model's performance with a low number of sampled queries answered by the oracle. Formally, \ac{AL} solves the following optimization problem:
\begin{equation} \label{eq:active_learning_problem}
    q^* = \argmax_{q \in \mathcal{Q}^{N_q}} \sum_{i=1}^{N_q} A(\theta, q_i)\,.
\end{equation}
The \ac{AIPP} problem in~\Cref{eq:ipp_problem} and \ac{AL} in~\Cref{eq:active_learning_problem} are closely related. In the \ac{AL} setup, the costs $C(q)=\lvert q \rvert$ of executing a sequence of queries $q$ is given by the number of queries $\lvert q \rvert$ in $q$. The optimization constraints of~\Cref{eq:active_learning_problem} are implicitly determined by $C(q) \leq N_q$. The \ac{AIPP} information criterion $I$ translates to $I(q) = \sum_{i=1}^{N_q} A(\theta, q_i)$ by substituting the \ac{AL} acquisition function as the mission objective. 

In the classical environmental monitoring or exploration problem setup, the environment $\xi$ is characterized by the set of available queries $\mathcal{Q}$, e.g., robot poses $q_i = \mathbf{x}_i \in \xi$. The feature space $\mathcal{F}$ encompasses the set of targets $\mathcal{L}$ establishing a correspondence between feature mappings \edit{$F: \xi \times \mathbb{R} \to \mathcal{F}$} and the model $\theta$. Further, we cannot directly quantify the true effect of a sequence of queries $q$ on the model performance $P(\theta)$ as this requires re-training on not-yet-acquired oracle-generated targets. Instead, similar to the belief update of $\Tilde{F}$ over the true feature mapping $F$ in~\Cref{eq:map_belief}, at each iteration $i$, \ac{AL} methods leverage already collected queries $\{q_1, \ldots, q_{i}\}$ and targets $\{O(q_1), \ldots, O(q_{i})\}$ to update the model $\theta$ and ensure maximally informed sequential query selection. We establish this theoretical connection between \ac{AIPP} and the \ac{AL} problem to show that the \ac{AIPP} methods introduced in \Cref{SS:supervised_learning,SS:deep_rl,SS:imitation_learning} can be viewed as methods for \ac{AL} in the context of embodied robotic applications. In the following, we identify the \textit{models}, \textit{queries}, \textit{oracles}, and \textit{acquisition functions} used in different learning-based approaches from an \ac{AL} perspective.

The \textit{model} could have any parametric or non-parametric form. As discussed in \Cref{SS:supervised_learning}, common models for learning a \edit{spatio(temporal) representation $F$} of a feature space $\mathcal{F}$ in an environment $\xi$ are non-parametric Gaussian processes or implicit neural representations to probabilistically model $O: \xi \to \mathcal{F}$. In this case, a \textit{query} $q_i$ is defined by a specific location or area in the environment $q_i = \mathbf{x}_i \in \xi$, and the \textit{oracle}'s answer $O(\mathbf{x}_i) = z_i$ is typically given by collecting a new potentially noisy measurement $z_i \in \mathcal{F}$ at the location $\mathbf{x}_i$ based on the sensor readings. As summarized in \Cref{T:mapping_metrics}, the performance metric $P$, i.e., evaluation metric, and thus \textit{acquisition function} $A$, i.e., information criterion $I$, varies with the task. Reductions in the covariance matrix trace of a Gaussian process model or model uncertainty of an implicit neural representation are popular acquisition functions used in exploration and monitoring tasks. \edit{Approaches leveraging a well-defined probabilistic map belief as a \textit{model} and maximising an information-theoretically motivated \textit{acquisition function} are often closely related to the theory of optimal experimental design~\citep{kiefer1959optimum}. Examples are \ac{AIPP} approaches that sequentially update Gaussian process map beliefs aiming to maximize its covariance trace reduction~\citep{marchant2014sequential, hitz2017adaptive, popovic2020informative, ott2023sboaippms, cao2023catnipp} as they are derived from A-optimal experiment design strategies}. Other supervised approaches learn the acquisition function $A$ with deep neural networks~\citep{hepp2018learn,bai2017toward,ly2019autonomous} or directly predict the next best query $q_{i+1}^*$, e.g., a view pose to reconstruct a 3D object~\citep{vasquez2021next,mendoza2020supervised}.

Similarly, \ac{RL}-based \ac{AIPP} approaches presented in~\Cref{SS:deep_rl} rely on a reward function $R: \mathcal{S} \times \mathcal{A} \times \mathcal{S} \to \mathbb{R}$ based on the current and next state and an action $a_i \in \mathcal{A}$ to learn a policy that predicts the action $a_{i+1}^* \in \mathcal{A}$ maximizing the return, i.e., the sum of rewards. According to the definitions of the return in~\Cref{eq:rl_return} and the \ac{AL} problem in~\Cref{eq:active_learning_problem}, the reward function in \ac{RL} acts as the acquisition function in \ac{AL}. In a \ac{POMDP} describing the \ac{AIPP} problem, the state $s$ can be seen as the current model $\theta$ trained on the history of collected queries $\{q_1, \ldots, q_{i}\}$ and targets $\{O(q_1), \ldots, O(q_{i})\}$. The queries $q_i=a_i \in \mathcal{A}$ correspond to the previously chosen actions $a_i$, e.g., the next measurement location, and the oracle-generated targets $O(q_{i})=z_i \in \mathcal{O}$ correspond to the observations, e.g., sensor readings at measurement locations.

The \ac{AIPP} approaches described so far assume static sensor models $p(z \,|\, F)$ to learn spatial environment models $\Tilde{F}$ in \Cref{eq:map_belief}. However, as environments $\xi$ are initially unknown, sensor models should be adapted to the robot environment to improve robot perception and, thus, the learned environment model $\Tilde{F}$. This is crucial for continuous robot deployment in varying domains and for deep learning-based sensor models as they are known to transfer poorly to unseen scenarios.

To address this problem, a line of recent research considers using \ac{AIPP} to actively gather data to improve a sensor model for robotic perception~\citep{ruckin2023informative,ruckin2022informative,blum2019active,zurbrugg2022embodied,chaplot2021seal}. In this problem setup, the information criterion $I$ in~\Cref{eq:ipp_problem} is defined in terms of how potential future raw sensor readings, e.g., RGB images at certain environment locations, and their associated oracle-generated targets, e.g., semantic segmentation labels, impact the sensor performance. To solve this problem, the key idea is to link the \ac{AIPP} objective to ideas from \ac{AL}.

To date, this problem has only been studied in the context of pixel-wise semantic segmentation from RGB images~\citep{ruckin2023informative,ruckin2022informative,blum2019active,zurbrugg2022embodied,chaplot2021seal}. These approaches can be divided into two categories depending on how they implement the oracle for labeling new data: self-supervised methods exploit pseudo labels rendered from an online-built semantic map as an oracle, without considering a human annotator, while fully supervised methods only exploit dense pixel-wise human annotations as an oracle. In the first category, self-supervised methods can be used to fuse 2D semantic predictions from different viewpoints into a semantic 3D map~\citep{zurbrugg2022embodied, chaplot2021seal}. After a mapping mission is completed, the semantic 3D map is used to automatically render consistent 2D pseudo labels for re-training. In this line of work, \citet{zurbrugg2022embodied} propose an \ac{AIPP} approach to plan viewpoints with high training data novelty for improved pseudo label quality. Similarly, \citet{chaplot2021seal} train an \ac{RL} agent to target low-confidence parts of a 3D map. Although these approaches do not require a human oracle~\cite{zurbrugg2022embodied, chaplot2021seal}, they still rely on large labeled pre-training datasets to generate high-quality pseudo labels in new scenes. In contrast, fully supervised methods are more applicable to varying domains and environments. \citet{blum2019active} locally plan paths of high training data novelty. \citet{ruckin2022informative} introduce a map-based planning framework supporting multiple acquisition functions~\citep{ruckin2023informative}. However, these methods~\citep{blum2019active, ruckin2022informative, ruckin2023informative} require considerable human labeling effort to improve robotic vision.

These works highlight the versatility of applying \ac{AL} methods to robotic learning through \ac{AIPP}. The combination of \ac{AL} and \ac{AIPP} methods demonstrates a principled approach to integrating learning-based components and classically used uncertainty quantification techniques for \ac{AL}. This integration represents a promising pathway towards robot deployment in more diverse environments with minimal human supervision. However, to harness the potential of \ac{AL} and \ac{AIPP}, future research must address multiple open problems. Existing uncertainty quantification methods for deep learning models are known to be overconfident~\citep{abdar2021review} and not well-calibrated in out-of-distribution scenarios. Further, fully and self-supervised approaches for sensor model improvement are limited in terms of human annotation effort and applicability to out-of-distribution environments, respectively. One possible direction is to take the strengths of both approaches and develop hybrid semi-supervised approaches~\citep{ruckin2023semi}. We may be able to deploy more versatile robots by combining \ac{AIPP} methods for environment monitoring and exploration (\Cref{SS:supervised_learning} to~\Cref{SS:imitation_learning}) with methods for targeted training data collection~\citep{ruckin2023informative,ruckin2022informative,blum2019active,zurbrugg2022embodied,chaplot2021seal}.

\begin{figure*}[!t]
    \centering
    \includegraphics[width=\textwidth]{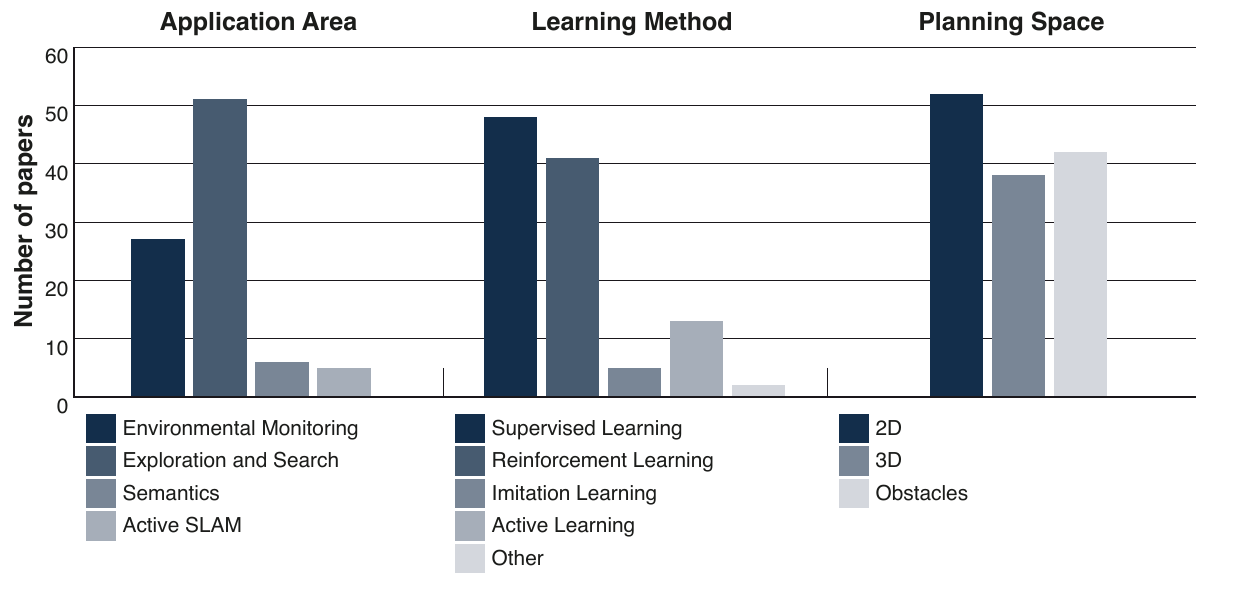}
\caption{Summary statistics for the papers reviewed in \Cref{S:applications}.} \label{F:histograms}
\end{figure*}


\section{Applications} \label{S:applications}

In this section, we categorize the \ac{AIPP} works surveyed in our paper according to their target applications in robotics. Our goal is to assess the practical applicability of existing learning-based methods while pinpointing emerging trends and recognizing potential limitations.

\Cref{T:applications} overviews \ac{AIPP} applications in robotics and describes their relevant practical aspects. Our taxonomy considers four broad application areas: (i) environmental monitoring; (ii) exploration and search; (iii) semantic scene understanding; and (iv) active \ac{SLAM}. For reference, in the last column, we also include the learning method: supervised learning (`SL'); reinforcement learning (`RL'); imitation learning (`IL'); and/or active learning (`AL'). \edit{In addition, in \Cref{F:histograms}, we provide visual summary statistics for our survey according to the application area, learning method, and planning space considered by each paper.}

In general, our survey illustrates that the works are broadly applied in terrestrial, aquatic, and aerial domains and on diverse robot platforms. \ac{AIPP} methods organically find usage in environmental monitoring scenarios to gather data about physical parameters, e.g., signal strength~\citep{hollinger2014sampling,denniston2023fast,bayerlein2021multi,denniston2023learned,yang2023intent,song2015trajectory}, temperature~\citep{meliou2007nonmyopic,ruckin2022adaptive,westheider2023multi}, or elevation~\citep{viseras2019deepig,zacchini2023informed}. In particular, \ac{AIPP} enables efficient data collection in large environments using sensors with limited coverage area, such as point-based sensors. For mapping, Gaussian processes are most commonly used to capture spatial correlations found in natural phenomena that are measured by a continuous variable. Several recent works use \ac{RL} to learn the best data-gathering actions~\citep{ruckin2022adaptive,cao2023catnipp,denniston2023learned}, including achieving coordinated monitoring with multi-robot teams~\citep{westheider2023multi,zhu2023multi,yanes2023deep,bayerlein2021multi}.

\ac{AIPP} strategies for exploration and search can be further broken down in terms of two distinct tasks: exploration of bounded unknown environments, i.e., outwards-facing scenarios, and active reconstruction of unknown objects or scenes, i.e., inwards-facing scenarios. The former necessarily accounts for possible obstacles in unknown regions and typically relies on depth-based sensing to construct volumetric maps as an environment is explored~\citep{zwecher2022integrating,lodel2022where,choudhury2018data,tzes2023graph,cao2023adriadne}. In contrast, most approaches for object reconstruction~\citep{vasquez2021next,lee2022uncertainty,jin2023neunbv,sunderhauf2023density,mendoza2020supervised} assume obstacle-free environments with a constrained hemispherical action space for a robot camera around the target object of interest. While NeRF-based methods are gaining popularity for this task~\citep{jin2023neunbv,sunderhauf2023density,lee2022uncertainty,yan2023active,zhang2023affordance,pan2022activenerf,pan2023views}, an open research challenge is applying them for exploration in more complex environments involving multiple objects and clutter.

Only a few works integrate semantic information or consider localization uncertainty in planning objectives for learning-based \ac{AIPP}. Several approaches~\citep{blum2019active, georgakis2022learning, zurbrugg2022embodied, ruckin2023informative, ruckin2022informative, ruckin2023semi} leverage semantics to improve deep learning-based sensor models via \ac{AL}. Although semantic scene understanding provides valuable cues for learning complex \ac{AIPP} behavior~\citep{schmid2022scexplorer}, a key development barrier is the lack of high-quality semantic datasets relevant to the domains of robotic \ac{AIPP} applications. The common assumption of perfect localization uncertainty represents a significant limitation to the applicability of existing methods, particularly in outdoor environments which may rely on imprecise \ac{GPS} signals~\citep{caley2019deep,hitz2017adaptive,westheider2023multi,choudhury2020adaptive,huttenrauch2019deep}. Many approaches also neglect sensor uncertainty and noise, especially in environmental monitoring scenarios where models for specialized sensors are scarce. This is a major drawback since imperfect sensing has been shown to have significant effects on the quality of collected data and resulting planning strategy~\citep{popovic2020informative,arora2019multi,zacchini2023informed}.

Our taxonomy reveals a gap in methods accounting for robot motion constraints while planning, with some methods assuming that the robot sensor can teleport between locations~\citep{jin2023neunbv,pan2022activenerf,sunderhauf2023density} or perform only incremental motions relative to its current position~\citep{westheider2023multi,viseras2021wildfire,bayerlein2021multi}. One open issue is the lack of guarantees on dynamic feasibility, which is essential for long-term path planning with agile, fast platforms such as small \acp{UAV}. Although there has been research on multi-robot missions~\citep{westheider2023multi,zhu2023multi,yang2023intent,gao2022cooperative,huttenrauch2019deep}, there are several areas that have not been widely studied, such as tasks involving multi-sensor fusion~\citep{ott2023sboaippms}, temporal changes in the environment~\citep{caley2019deep,marchant2014sequential}, and combining multiple problems simultaneously, e.g., semantic scene understanding and active \ac{SLAM}.

\begin{landscape}
\footnotesize

\begin{table}[!t]
  \centering
  \footnotesize

  {\renewcommand{\arraystretch}{1.019}
  \setlength{\tabcolsep}{4.5pt}
  \footnotesize

  \begin{threeparttable}
  
  \begin{tabular} {@{}c p{0.03\textwidth} >{\raggedright}p{0.25\textwidth} p{0.11\textwidth} >{\raggedright}p{0.23\textwidth} >{\centering}p{0.07\textwidth} p{0.06\textwidth} >{\raggedright}p{0.18\textwidth} p{0.15\textwidth} p{0.08\textwidth}@{}}
    \toprule
     & \multicolumn{1}{c}{} & \multicolumn{1}{l}{\multirow{2}{*}{Domain}} & \multicolumn{1}{l}{\multirow{2}{*}{Robot platforms}} & \multicolumn{1}{l}{\multirow{2}{*}{Sensors}} & \multicolumn{1}{l}{\multirow{2}{*}{Obstacles?}} & \multicolumn{1}{l}{\multirow{2}{*}{\parbox{0.06\textwidth}{Planning \\ space}}} & \multicolumn{1}{l}{\multirow{2}{*}{Sources of uncertainty}} & \multicolumn{1}{l}{\multirow{2}{*}{\parbox{0.15\textwidth}{Motion \\ constraints}}} & \multicolumn{1}{l}{\multirow{2}{*}{\parbox{0.08\textwidth}{Learning \\ method}}} \\ \\ \midrule
     
     \multirow{35}{*}{\STAB{\rotatebox[origin=c]{90}{Environmental monitoring}}}
     & \citep{velasco2020adaptive}  & Air quality monitoring & UAV & Point-based gas sensor & \xmark  & 2D  & Mapping & - & - \\
     & \citep{choi2021adaptive}  & Air quality monitoring & UGV & Point-based sensor & \xmark  & 2D  & Mapping & - & SL, RL  \\
     & \citep{viseras2019deepig}  & Elevation mapping & UAV team & Ultrasound height sensor & \cmark  & 2D & Mapping & - & SL, RL \\
     & \citep{zeng2022deep}  & Fruit monitoring  & Robot arm & RGB-D camera & \cmark  & 3D  & Mapping & 
     ROS MoveIt$^*$ & RL \\
     & \citep{menon2023nbvsc}  & Fruit monitoring  & Robot arm & RGB-D camera & \cmark  & 3D  & Mapping & ROS MoveIt$^*$ & SL \\
     & \citep{denniston2023learned}  & Intensity monitoring & AUV/UGV/UAV & Generic camera, point-based sensor  & \xmark & 2D/3D  & Mapping & - & SL, RL \\
     & \citep{cao2023catnipp}  & Intensity monitoring  & UGV & Point-based sensor & \xmark  & 2D  & Mapping & - & SL, RL \\
     & \citep{yang2023intent}  & Intensity monitoring  & UGV team & Point-based sensor & \xmark  & 2D & Mapping & - & SL, RL \\
     & \citep{song2015trajectory} & Intensity monitoring & UAV & Point-based sensor & \xmark & 2D  & Mapping, sensing & Nonholonomic constraints & SL \\
     & \citep{hitz2017adaptive}  & Lake bacteria monitoring  & AUV & Point sensor for phycoerythrin fluorescence & \xmark  & 3D & Mapping & Spline path & SL \\
     & \citep{duecker2021embedded} & Ocean pollution monitoring & AUV & Pressure sensor, general water characterization & \xmark & 2D & Mapping & Spline path & RL \\
     & \citep{popovic2020informative}  & Precision agriculture  & UAV & Generic camera & \xmark  & 3D  & Mapping, sensing, semantics & Spline path & SL \\
     & \citep{ruckin2022adaptive}  & Precision agriculture  & UAV & Generic camera & \xmark  & 3D & Mapping, sensing & - & SL, RL \\
     & \citep{westheider2023multi}  & Precision agriculture  & UAV team & Generic camera & \xmark  & 3D & Mapping, sensing & - & RL \\
     & \citep{arora2019multi}  & Scientific data gathering  & UGV & Generic camera, ultraviolet light source, laser scanner  & \xmark & 2D  & Mapping, sensing & Omnidirectional drive & RL \\
     & \citep{hollinger2014sampling} & Signal strength monitoring & AUV & Point-based sensor & \xmark & 2D & Mapping & - & SL \\
     & \citep{denniston2023fast}  & Signal strength monitoring & AUV/UGV & Generic camera, point-based sensor & \cmark  & 2D & Mapping, path executability & - & RL \\ 
     & \citep{wei2020informative}  & Signal strength monitoring  & UGV & Point-based sensor & \xmark  & 2D & Mapping & - & SL, RL \\
     & \citep{caley2019deep}  & Spatiotemporal field monitoring  & AUV/UGV & Point-based sensor & \xmark  & 2D & Mapping & - & SL \\
     & \citep{marchant2014sequential}  & Spatiotemporal field monitoring  & UGV & Point-based sensor & \xmark  & 2D  & Mapping & Spline path & SL, RL \\
     & \citep{tzes2023graph}  & Target tracking and localization & UGV team & RGB-D sensor & \xmark  & 2D  & Mapping, sensing & - & IL \\
     & \citep{meliou2007nonmyopic}  & Temperature monitoring  & Sensor network & Generic & \xmark  & 2D  & Mapping & - & SL \\
     & \citep{vivaldini2019uav}  & Tree disease classification & UAV & RGB camera & \xmark  & 2D  & Mapping, semantics & - & SL \\
     & \citep{zacchini2023informed} & Underwater mapping & AUV & Sonar  & \xmark & 3D  & Mapping, sensing & Dubins vehicle model & SL \\
     & \citep{yanes2023deep}  & Water quality monitoring & AUV team & Water quality sensor & \xmark  & 2D  & Mapping & - & SL, RL \\
     & \citep{viseras2021wildfire}  & Wildfire monitoring  & UAV team & Thermal camera & \xmark  & 2D & -  & - & RL \\
     & \citep{bayerlein2021multi}  & Wireless data harvesting  & UAV team & Signal receptor & \cmark  & 2D & Signal emissions & - & RL \\
     \midrule
     
     \multirow{10}{*}{\STAB{\rotatebox[origin=c]{90}{Exploration and search}}}
     & \citep{dai2022camera} & Camera view planning & UGV & RGB-D camera & \cmark & 3D  & - & - & IL \\
     & \citep{yang2023learning}  & Exploration and landmark localization  & UAV & Depth sensor & \xmark  & 2D/3D  & Mapping & - & RL \\
     & \citep{zhu2023multi}  & Exploration and target search  & UGV team & Depth sensor & \cmark  & 2D  & - & - & RL \\
     & \citep{choudhury2018data}  & Exploration, search-based planning  & UGV/UAV & RGB-D camera & \cmark  & 2D  & Mapping & - & IL \\
     & \citep{lodel2022where}  & Floor plan exploration  & UAV & Laser & \cmark  & 2D & Mapping, sensing & Unicycle model & RL \\
     & \citep{liu2023learning}  & Indoor exploration  & UGV & Depth sensor & \cmark  & 2D & -  & - & RL, IL \\
     & \citep{zwecher2022integrating}  & Indoor exploration  & UGV & Depth sensor & \cmark  & 2D & Map prediction & - & SL, RL \\
     & \citep{bai2017toward}  & Indoor exploration  & UGV & Depth sensor & \cmark  & 2D & Mapping, sensing & - & SL \\
     & \citep{tao2023seer}  & Indoor exploration & UAV & Depth sensor & \cmark  & 3D  & Mapping, sensing & Spline path & SL \\
     & \citep{cao2023adriadne}  & Indoor exploration & UGV & Depth sensor & \cmark  & 2D  & Mapping, sensing & Local planner$^*$~\citep{cao2022autonomous} & RL \\
     \bottomrule
  \end{tabular}

    \begin{tablenotes}
      \item $*$ Motion constraints incorporated only during plan execution.
    \end{tablenotes}
  \end{threeparttable}}

\caption{\label{T:applications} Taxonomy of AIPP applications in robotics.}

\end{table}
\newpage \strut
\newpage \strut
\footnotesize
  {\renewcommand{\arraystretch}{1.019}
  \setlength{\tabcolsep}{4.5pt}
  \begin{tabular} {@{}c p{0.03\textwidth} >{\raggedright}p{0.25\textwidth} p{0.11\textwidth} >{\raggedright}p{0.23\textwidth} >{\centering}p{0.07\textwidth} p{0.06\textwidth} >{\raggedright}p{0.18\textwidth} p{0.15\textwidth} p{0.08\textwidth}@{}}
    \toprule
     & \multicolumn{1}{c}{} & \multicolumn{1}{l}{\multirow{2}{*}{Domain}} & \multicolumn{1}{l}{\multirow{2}{*}{Robot platforms}} & \multicolumn{1}{l}{\multirow{2}{*}{Sensors}} & \multicolumn{1}{l}{\multirow{2}{*}{Obstacles?}} & \multicolumn{1}{l}{\multirow{2}{*}{\parbox{0.06\textwidth}{Planning \\ space}}} & \multicolumn{1}{l}{\multirow{2}{*}{Sources of uncertainty}} & \multicolumn{1}{l}{\multirow{2}{*}{\parbox{0.15\textwidth}{Motion \\ constraints}}} & \multicolumn{1}{l}{\multirow{2}{*}{\parbox{0.08\textwidth}{Learning \\ method}}} \\ \\ \midrule
     
     \multirow{46}{*}{\STAB{\rotatebox[origin=c]{90}{Exploration and search}}}
     & \citep{niroui2019deep}  & Indoor exploration & UGV & Depth sensor & \cmark  & 2D  & Mapping, sensing & ROS move\_base$^*$ & RL \\
     & \citep{chen2019self}  & Indoor exploration  & UGV & Depth sensor & \cmark  & 2D/3D  & Mapping, sensing & - & RL \\
     & \citep{shrestha2019learned} & Indoor exploration & UGV & Laser scanner & \cmark & 2D & Mapping & - & SL \\
     & \citep{li2023learning}  & Indoor exploration  & UGV & RGB-D camera & \cmark  & 2D & - & Local planner using Fast Marching Method & SL \\
     & \citep{schmid2022fast} & Indoor exploration & UGV & RGB-D camera & \cmark  & 2D  & Mapping, sensing & - & SL \\
     & \citep{schmid2022scexplorer}  & Indoor exploration & UAV & RGB-D camera & \cmark  & 3D  & Mapping, sensing, scene completion & Velocity ramp model & SL \\
     & \citep{georgakis2022upen} & Indoor exploration, point-goal navigation & UGV/UAV & Depth sensor & \cmark & 2D & Mapping, sensing & - & SL \\
     & \citep{ramakrishnan2020occupancy} & Indoor exploration & UGV/UAV & RGB-D camera & \cmark & 2D & Mapping, sensing & - & SL \\
     & \citep{gazani2023bag}  & Infrastructure inspection  & UAV & RGB-D camera & \xmark  & 3D & -  & - & RL \\
     & \citep{kumar2022graph}  & Interactive exploration  & Robot arm & RGB-D camera & \xmark  & 3D & -  & - & RL \\
     & \citep{chen2022learning}  & Multi-object navigation  & UGV & RGB-D camera & \cmark  & 2D  & - & Revolute joint camera & RL \\
     & \citep{novkovic2020object}  & Object finding  & Robot arm & RGB-D camera & \cmark  & 3D & Mapping, sensing  & - & RL \\
     & \citep{best2019dec}  & Object recognition, team orienteering  & UGV team & 2D Laser & \cmark  & 2D  & Mapping, sensing & - & RL \\
     & \citep{mendoza2020supervised}  & Object reconstruction  & Any platform & Depth sensor & \xmark  & 3D & -  & - & SL \\
     & \citep{dhami2023pred}  & Object reconstruction & Any platform & Depth sensor & \cmark  & 3D & -  & - & SL \\
     & \citep{vasquez2021next}  & Object reconstruction & Any platform & Depth sensor & \xmark  & 3D & Mapping & - & SL \\
     & \citep{pan2023views}  & Object reconstruction & Any platform & RGB camera & \xmark  & 3D  & - & - & SL \\ 
     & \citep{pan2022activenerf}  & Object reconstruction & Any platform & RGB camera & \xmark  & 3D  & Sensing & - & SL, AL \\ 
     & \citep{zhan2022activermap}  & Object reconstruction & Any platform & RGB camera & \xmark  & 3D  & Sensing & - & SL, AL \\ 
     & \citep{sunderhauf2023density}  & Object reconstruction & Any platform & RGB camera & \xmark  & 3D  & Sensing & - & SL, AL \\ 
     & \citep{jin2023neunbv}  & Object reconstruction & Any platform & RGB camera & \xmark  & 3D & Sensing & - & SL, AL \\ 
     & \citep{sukkar2019multi}  & Object reconstruction  & Robot arm &  RGB-D camera  & \cmark  & 3D  & Mapping, sensing & - & RL \\
     & \citep{lee2022uncertainty}  & Object reconstruction & Any platform & RGB-D camera & \xmark  & 3D  & Sensing & - & SL, AL \\ 
     & \citep{ran2022neurar}  & Object reconstruction & Any platform & RGB-D camera & \xmark & 3D  & Sensing & - & SL, AL \\ 
     & \citep{yan2023active}  & Object reconstruction & Any platform & RGB-D camera & \cmark  & 3D & Sensing & - & SL, AL \\
     & \citep{dhami2023pred}  & Object reconstruction  & UAV & Depth sensor & \cmark  & 3D  & - & ROS MoveIt$^*$ & SL \\
     & \citep{pan2023oneshot}  & Object surface reconstruction  & Any platform & RGB-D camera & \xmark  & 3D  & Sensing & ROS MoveIt$^*$ & SL \\
     & \citep{wang2023spatiotemporal}  & Persistent target monitoring  & Robot team (UAV/UGV) & Generic camera & \xmark  & 2D & Target position & - & SL, RL \\
     & \citep{ott2023sboaippms} & Rover exploration & UGV & Drill, mass spectrometer & \cmark & 2D & Mapping, sensing & - & RL \\
     & \citep{choudhury2020adaptive}  & Search and rescue  & Any platform & Categorical survivor detection & \cmark  & 2D  & Mapping, sensing & - & RL\\
     & \citep{singh2009nonmyopic} & Search and rescue & Any platform & Optical sensor & \xmark & 2D & Sensing & - & RL \\
     & \citep{lim2016adaptive}  & Search tasks, e.g., grasping pose or target search & Any platform & Binary measurements & \xmark  & 3D  & Sensing & - & - \\
     & \citep{reinhart2020learning}  & Subterranean exploration & UAV & Depth sensor & \cmark  & 3D & -  & - & IL \\
     & \citep{qie2019joint}  & Target assignment  & UAV team & Not specified & \xmark  & 2D & -  & - & RL \\
     & \citep{gao2022cooperative}  & Target assignment  & UAV team & Not specified & \xmark & 2D & Mapping, sensing & - & RL \\
     & \citep{huttenrauch2019deep}  & Target tracking  & UAV team & Laser  & \xmark  & 2D & Mapping & Unicycle model & RL \\
     & \citep{ye2018active} & Targeted object search & UGV & RGB camera & \cmark  & 2D & Object recognition & - & RL \\
     & \citep{saroya2020online}  & Tunnel exploration  & UGV & Depth sensor & \cmark  & 2D & -  & - & SL \\
     & \citep{ly2019autonomous}  & Urban exploration, surveillance  & UGV/UAV & Depth sensor & \cmark  & 2D & - & - & SL \\
     & \citep{hepp2018learn} & Urban exploration & UAV & Depth sensor & \cmark  & 3D  & Mapping, sensing & - & SL \\      
     & \citep{zhang2023affordance}  & View planning for grasping & Any platform & RGB-D camera & \xmark & 3D  & -  & - & SL, AL \\ 
     \bottomrule
     
\end{tabular}}

\strut\newpage

\strut\newpage

\footnotesize
  {\renewcommand{\arraystretch}{1.1019}
  \setlength{\tabcolsep}{4.5pt}
  
  \begin{tabular} {@{}c p{0.03\textwidth} >{\raggedright}p{0.25\textwidth} p{0.11\textwidth} >{\raggedright}p{0.23\textwidth} >{\centering}p{0.07\textwidth} p{0.06\textwidth} >{\raggedright}p{0.18\textwidth} p{0.15\textwidth} p{0.08\textwidth}@{}}
    \toprule
     & \multicolumn{1}{c}{} & \multicolumn{1}{l}{\multirow{2}{*}{Domain}} & \multicolumn{1}{l}{\multirow{2}{*}{Robot platforms}} & \multicolumn{1}{l}{\multirow{2}{*}{Sensors}} & \multicolumn{1}{l}{\multirow{2}{*}{Obstacles?}} & \multicolumn{1}{l}{\multirow{2}{*}{\parbox{0.06\textwidth}{Planning \\ space}}} & \multicolumn{1}{l}{\multirow{2}{*}{Sources of uncertainty}} & \multicolumn{1}{l}{\multirow{2}{*}{\parbox{0.15\textwidth}{Motion \\ constraints}}} & \multicolumn{1}{l}{\multirow{2}{*}{\parbox{0.08\textwidth}{Learning \\ method}}} \\ \\ \midrule

     \multirow{8}{*}{\STAB{\rotatebox[origin=c]{90}{Semantics}}}
     & \citep{blum2019active}  & Land cover mapping & UAV & RGB camera & \xmark  & 2D  & Sensing & - & AL \\
     & \citep{georgakis2022learning}  & Indoor goal navigation  & UGV & RGB-D camera & \cmark  & 2D & Mapping, sensing & - & AL \\
     & \citep{zurbrugg2022embodied} & Indoor semantic mapping & UGV & Laser, RGB-D camera & \cmark  & 3D & Mapping, sensing, localization & - & AL \\
     & \citep{ruckin2023informative}  & Industrial and urban mapping, land cover analysis & UAV & RGB camera & \xmark  & 2D & Mapping, sensing & - \\      
     & \citep{ruckin2022informative}  & Urban mapping & UAV & RGB camera & \xmark  & 2D & Mapping, sensing & - & AL \\
     & \citep{ruckin2023semi}  & Urban mapping & UAV & RGB camera & \xmark  & 2D & Mapping, sensing & - & AL \\
     \midrule
     
     \multirow{9}{*}{\STAB{\rotatebox[origin=c]{90}{Active SLAM}}}
     & \citep{hanlon2023active}  & Active visual localization  & UGV & RGB-D camera & \cmark  & 3D & Localization & - & SL \\
     & \citep{popovic2020informativelocunc}  & Indoor temperature monitoring & UGV/UAV & Laser, temperature sensor & \cmark  & 2D/3D & Mapping, sensing, localization & - & SL \\
     & \citep{chen2020autonomous}  & Exploration  & UGV & Lidar & \xmark & 2D & Mapping, sensing, localization & - & RL \\
     & \citep{chen2021zeroshot}  & Exploration  & UGV & Lidar & \cmark  & 2D  & Mapping, sensing, localization & - & RL \\ 
     & \citep{bartolomei2021semantic} & Goal-based navigation & UAV & RGB-D camera & \cmark & 3D & Mapping, sensing, localization & Spline path & RL \\
     \bottomrule
     
\end{tabular}}
  
\end{landscape}

\section{Challenges and Future Directions} \label{S:discussion}
This section outlines the open challenges related to deploying learning-based \ac{AIPP} in robotics settings and avenues for future work to overcome them.

\subsection{Generalizability} \label{SS:generalizability}

Most of the \ac{AIPP} methods surveyed were tested in the same or similar environments as those seen during training. The generalizability of deep learning-based systems is especially important because the neural networks that are often used in \ac{AIPP} frameworks~\citep{choudhury2018data, shrestha2019learned, tao2023seer, schmid2022scexplorer, li2020graph, dai2022camera} are known to produce overconfident wrong predictions in out-of-distribution scenarios~\citep{abdar2021review}. Future work should address these inherent generalization limitations to improve robustness. We identified one possible way to deal with deep learning-based vision uncertainties in \Cref{SS:active_learning} by combining uncertainty-aware deep learning and active learning approaches with \ac{AIPP}~\citep{chaplot2021seal, zurbrugg2022embodied, ruckin2023informative}. Similarly, \ac{RL}-based \ac{AIPP} systems are increasingly popular~\citep{cao2023catnipp, yang2023intent, cao2023adriadne, chen2021zeroshot, wei2020informative, choi2021adaptive}. However, \ac{RL} approaches degrade in performance when deployed in unseen simulated environments or real-world scenarios~\citep{bousmalis2018using, zhao2020sim, zhu2020ingredients}. Future work on \ac{RL}-based \ac{AIPP} methods should focus on evaluating the learned policy in more diverse environments and in the real world. Further, new methods are needed to improve the generalization of \ac{RL} for \ac{AIPP}~\citep{kirk2023survey}. Other subfields of \ac{RL}, such as meta-reinforcement learning~\citep{gupta2018meta}, could serve as inspiration for enabling efficient online policy adaption to unseen environments. Data augmentation and domain randomization during training could improve test-time policy robustness~\citep{sadeghicad2rl}.

\subsection{Handling Localization Uncertainty} \label{SS:localization uncertainty}

Effectively handling robot localization uncertainty is a key aspect in advancing learning-based \ac{AIPP} in robotics. Localization uncertainties arise from various sources, including sensor noise, environmental changes, and the inherent limitations of localization technologies. However, most existing approaches assume perfect knowledge of the robot pose, both during planning and plan execution, which limits their practical applicability. Some works in active \ac{SLAM} account for localization uncertainty via belief space planning~\citep{chen2020autonomous,chen2021zeroshot} or modifying the planning objective~\citep{popovic2020informativelocunc} using graph-based representations of the robot state. Incorporating such probabilistic methods into learning-based \ac{AIPP} methods and exploring transfer learning techniques~\citep{chen2021zeroshot,genc2020zeroshot} can enhance robustness.

\subsection{Temporal Changes and Long-Term Planning} \label{SS:temporal}

Addressing temporal changes and ensuring robust long-term planning in learning-based \ac{AIPP} poses a critical challenge. Temporal changes, such as variations in environmental conditions, seasonal shifts, or the presence of dynamic obstacles, require \ac{AIPP} methods to not only adapt to the immediate surroundings but also plan for sustained effectiveness over long durations. This involves developing strategies that consider the evolution of the environment and adjust the planned paths online. Whereas several works study spatiotemporal monitoring~\citep{caley2019deep,marchant2014sequential,wang2023spatiotemporal} and dynamic target tracking~\citep{tzes2023graph,huttenrauch2019deep}, there is a need for new approaches that can handle multiple moving objects in 3D environments. We also highlight the importance of developing such methods for agile platforms such as small \ac{UAV} platforms while also accounting for their physical motion constraints, given that they often operate in dynamic and changing landscapes.

\subsection{Heterogeneity} \label{SS:heterogeneity}

One challenge is the development of learning-based \ac{AIPP} methods that exploit the benefits of heterogeneous robot teams or heterogeneous sensor modalities within a single robot. Leveraging heterogeneity in robotic systems has a high potential to improve sensing and monitoring performance. For instance, a tandem robotic system can combine the agility of a small \ac{UAV} and the high-payload, detailed inspection capabilities of a \ac{UGV}. A robot equipped with different sensors, e.g., a drill and a mass spectrometer~\citep{ott2023sboaippms}, can measure different physical parameters during a single mission. To fully exploit these capabilities, we recommend investigating principled ways of incorporating heterogeneity, also considering collaboration aspects for effective decision-making.

\subsection{Standardized Evaluation Methods} \label{SS:standardized_evaluation}

Another key challenge is the lack of standardized simulation environments, evaluation metrics, and benchmarks to assess new methods in \ac{AIPP}. As highlighted in~\Cref{S:background}, most existing works rely on custom-made procedures and baselines for evaluation. While these reveal relative trends in performance, they do not ensure reproducible results across different approaches. Moreover, our survey catalogue\footnote{\url{https://dmar-bonn.github.io/aipp-survey/}} indicates that very few works have publicly available datasets, documentation, and open-source code. In future work, it is essential to establish common evaluation pipelines to promote reproducibility and comparability. We recommend constructing such a framework in a modular fashion, allowing easy modification of various aspects, including the application scenario (robot platform, sensor setup, and environment) and algorithmic components (mapping method, planning objective, and planning algorithm). We believe such a development would also help guide future work towards more general approaches, rather than constrained application-specific methods, as well as foster stronger collaboration within the \ac{AIPP} community.

\section{Conclusion} \label{S:conclusion}
In this survey paper, we analyzed the current research on \ac{AIPP} in robotic applications, focusing especially on methods that exploit recent advances in robot learning. We introduced a unified \edit{mathematical problem definition} for addressing \edit{tasks} in learning-based \ac{AIPP}. Based on this formulation, we provided two complementary taxonomies from the perspectives of (i) learning algorithms and (ii) robotic applications. We provided a structured outline of the research landscape and highlighted synergies and trends. Finally, we discussed the remaining challenges for learning-based \ac{AIPP} in robotics and outlined promising research directions.  

\section{Funding Sources}

This work was funded by the Deutsche Forschungsgemeinschaft (DFG, German Research Foundation) under Germany's Excellence Strategy - EXC 2070 – 390732324.

We would like to thank Jonas Schaap for helping with the infographics design.

\bibliographystyle{IEEEtranN}
\footnotesize
\bibliography{2023-survey}

\begin{thebibliography}{146}
\providecommand{\natexlab}[1]{#1}
\providecommand{\url}[1]{#1}
\csname url@samestyle\endcsname
\providecommand{\newblock}{\relax}
\providecommand{\bibinfo}[2]{#2}
\providecommand{\BIBentrySTDinterwordspacing}{\spaceskip=0pt\relax}
\providecommand{\BIBentryALTinterwordstretchfactor}{4}
\providecommand{\BIBentryALTinterwordspacing}{\spaceskip=\fontdimen2\font plus
\BIBentryALTinterwordstretchfactor\fontdimen3\font minus \fontdimen4\font\relax}
\providecommand{\BIBforeignlanguage}[2]{{%
\expandafter\ifx\csname l@#1\endcsname\relax
\typeout{** WARNING: IEEEtranN.bst: No hyphenation pattern has been}%
\typeout{** loaded for the language `#1'. Using the pattern for}%
\typeout{** the default language instead.}%
\else
\language=\csname l@#1\endcsname
\fi
#2}}
\providecommand{\BIBdecl}{\relax}
\BIBdecl

\bibitem[Rayas~Fernández et~al.(2022)Rayas~Fernández, Denniston, Caron, and Sukhatme]{rayas2022informative}
I.~M. Rayas~Fernández, C.~E. Denniston, D.~A. Caron, and G.~S. Sukhatme, ``{Informative Path Planning to Estimate Quantiles for Environmental Analysis},'' \emph{IEEE Robotics and Automation Letters (RA-L)}, vol.~7, no.~4, pp. 10\,280--10\,287, 2022.

\bibitem[Cao et~al.(2023{\natexlab{a}})Cao, Wang, Vashisth, Fan, and Sartoretti]{cao2023catnipp}
Y.~Cao, Y.~Wang, A.~Vashisth, H.~Fan, and G.~A. Sartoretti, ``{CAtNIPP: Context-Aware Attention-based Network for Informative Path Planning},'' in \emph{Proc.~of the Conf.~on Robot Learning (CoRL)}, 2023.

\bibitem[Popović et~al.(2020)Popović, Vidal-Calleja, Hitz, Chung, Sa, Siegwart, and Nieto]{popovic2020informative}
M.~Popović, T.~Vidal-Calleja, G.~Hitz, J.~J. Chung, I.~Sa, R.~Siegwart, and J.~Nieto, ``{An informative path planning framework for UAV-based terrain monitoring},'' \emph{Autonomous Robots}, vol.~44, pp. 889--911, 2020.

\bibitem[Denniston et~al.(2023{\natexlab{a}})Denniston, Peltzer, Ott, Moon, Kim, Sukhatme, Kochenderfer, Schwager, and Agha-mohammadi]{denniston2023fast}
C.~E. Denniston, O.~Peltzer, J.~Ott, S.~Moon, S.-K. Kim, G.~S. Sukhatme, M.~J. Kochenderfer, M.~Schwager, and A.-a. Agha-mohammadi, ``{Fast and Scalable Signal Inference for Active Robotic Source Seeking},'' in \emph{Proc.~of the IEEE Intl.~Conf.~on Robotics \& Automation (ICRA)}, 2023.

\bibitem[R{\"u}ckin et~al.(2022{\natexlab{a}})R{\"u}ckin, Jin, and Popović]{ruckin2022adaptive}
J.~R{\"u}ckin, L.~Jin, and M.~Popović, ``{Adaptive Informative Path Planning Using Deep Reinforcement Learning for UAV-based Active Sensing},'' in \emph{Proc.~of the IEEE Intl.~Conf.~on Robotics \& Automation (ICRA)}, 2022.

\bibitem[Viseras and Garcia(2019)]{viseras2019deepig}
A.~Viseras and R.~Garcia, ``{DeepIG: Multi-robot information gathering with deep reinforcement learning},'' \emph{IEEE Robotics and Automation Letters (RA-L)}, vol.~4, no.~3, pp. 3059--3066, 2019.

\bibitem[Choudhury et~al.(2020)Choudhury, Gruver, and Kochenderfer]{choudhury2020adaptive}
S.~Choudhury, N.~Gruver, and M.~J. Kochenderfer, ``{Adaptive Informative Path Planning with Multimodal Sensing},'' in \emph{International Conference on Automated Planning and Scheduling (ICAPS)}, 2020.

\bibitem[Singh et~al.(2009)Singh, Krause, and Kaiser]{singh2009nonmyopic}
A.~Singh, A.~Krause, and W.~J. Kaiser, ``{Nonmyopic Adaptive Informative Path Planning for Multiple Robots},'' in \emph{Proc.~of the Intl.~Conf.~on Artificial Intelligence (IJCAI)}, 2009.

\bibitem[Galceran and Carreras(2013)]{galceran2013survey}
E.~Galceran and M.~Carreras, ``{A Survey on Coverage Path Planning for Robotics},'' \emph{Journal on Robotics and Autonomous Systems (RAS)}, vol.~61, no.~12, pp. 1258--1276, 2013.

\bibitem[Tan et~al.(2021)Tan, Mohd-Mokhtar, and Arshad]{tan2021comprehensive}
C.~S. Tan, R.~Mohd-Mokhtar, and M.~R. Arshad, ``{A Comprehensive Review of Coverage Path Planning in Robotics Using Classical and Heuristic Algorithms},'' \emph{IEEE Access}, vol.~9, pp. 119\,310--119\,342, 2021.

\bibitem[Bai et~al.(2021)Bai, Shan, Chen, Liu, and Englot]{bai2021information}
S.~Bai, T.~Shan, F.~Chen, L.~Liu, and B.~Englot, ``{Information-Driven Path Planning},'' \emph{Current Robotics Reports}, no.~2, pp. 177--188, 2021.

\bibitem[Maboudi et~al.(2023)Maboudi, Homaei, Song, Malihi, Saadatseresht, and Gerke]{maboudi2023review}
M.~Maboudi, M.~Homaei, S.~Song, S.~Malihi, M.~Saadatseresht, and M.~Gerke, ``{A Review on Viewpoints and Path Planning for UAV-Based 3D Reconstruction},'' \emph{IEEE Journal of Selected Topics in Applied Earth Observations and Remote Sensing}, 2023.

\bibitem[Lluvia et~al.(2021)Lluvia, Lazkano, and Ansuategi]{lluvia2021active}
I.~Lluvia, E.~Lazkano, and A.~Ansuategi, ``{Active Mapping and Robot Exploration: A Survey},'' \emph{Sensors}, vol.~21, no.~7, 2021.

\bibitem[Sung et~al.(2023)Sung, Das, and Tokekar]{sung2023decision}
Y.~Sung, J.~Das, and P.~Tokekar, ``{Decision-Theoretic Approaches for Robotic Environmental Monitoring -- A Survey},'' \emph{arXiv preprint arXiv:2308.02698}, 2023.

\bibitem[Taylor et~al.(2021)Taylor, Berrueta, and Murphey]{taylor2021active}
A.~T. Taylor, T.~A. Berrueta, and T.~D. Murphey, ``{Active learning in robotics: A review of control principles},'' \emph{Mechatronics}, vol.~77, p. 102576, 2021.

\bibitem[Aniceto and Vivaldini(2022)]{dos2022review}
M.~Aniceto and K.~C.~T. Vivaldini, ``{A Review of the Informative Path Planning, Autonomous Exploration and Route Planning Using UAV in Environment Monitoring},'' in \emph{Intl. Conf.~on Computational Science and Computational Intelligence (CSCI)}, 2022.

\bibitem[Argall et~al.(2009)Argall, Chernova, Veloso, and Browning]{argall2009survey}
B.~D. Argall, S.~Chernova, M.~Veloso, and B.~Browning, ``A survey of robot learning from demonstration,'' \emph{Robotics and autonomous systems}, vol.~57, no.~5, pp. 469--483, 2009.

\bibitem[Garaffa et~al.(2023)Garaffa, Basso, Konzen, and de~Freitas]{garaffa2023reinforcement}
L.~C. Garaffa, M.~Basso, A.~A. Konzen, and E.~P. de~Freitas, ``{Reinforcement Learning for Mobile Robotics Exploration: A Survey},'' \emph{{IEEE Transactions on Neural Networks and Learning Systems}}, vol.~34, no.~8, pp. 3796--3810, 2023.

\bibitem[Lauri et~al.(2022)Lauri, Hsu, and Pajarinen]{lauri2022partially}
M.~Lauri, D.~Hsu, and J.~Pajarinen, ``Partially observable {M}arkov decision processes in robotics: A survey,'' \emph{IEEE Trans.~on Robotics (TRO)}, vol.~39, no.~1, pp. 21--40, 2022.

\bibitem[Mukherjee et~al.(2022)Mukherjee, Gupta, Chang, and Najjaran]{mukherjee2022survey}
D.~Mukherjee, K.~Gupta, L.~H. Chang, and H.~Najjaran, ``A survey of robot learning strategies for human-robot collaboration in industrial settings,'' \emph{Robotics and Computer-Integrated Manufacturing}, vol.~73, p. 102231, 2022.

\bibitem[Chen(2011)]{chen2011kalman}
S.~Chen, ``Kalman filter for robot vision: a survey,'' \emph{IEEE Transactions on Industrial Electronics}, vol.~59, no.~11, pp. 4409--4420, 2011.

\bibitem[Elfes(1989)]{elfes1989using}
A.~Elfes, ``Using occupancy grids for mobile robot perception and navigation,'' \emph{Computer}, vol.~22, no.~6, pp. 46--57, 1989.

\bibitem[Rasmussen and Williams(2006)]{rasmussen2006gaussian}
C.~E. Rasmussen and C.~K.~I. Williams, \emph{{Gaussian Processes for Machine Learning}}.\hskip 1em plus 0.5em minus 0.4em\relax MIT Press, 2006.

\bibitem[Mildenhall et~al.(2021)Mildenhall, Srinivasan, Tancik, Barron, Ramamoorthi, and Ng]{mildenhall2021nerf}
B.~Mildenhall, P.~P. Srinivasan, M.~Tancik, J.~T. Barron, R.~Ramamoorthi, and R.~Ng, ``{Nerf: Representing scenes as neural radiance fields for view synthesis},'' \emph{Communications of the ACM}, vol.~65, no.~1, pp. 99--106, 2021.

\bibitem[Placed et~al.(2023)Placed, Strader, Carrillo, Atanasov, Indelman, Carlone, and Castellanos]{placed2023survey}
J.~A. Placed, J.~Strader, H.~Carrillo, N.~Atanasov, V.~Indelman, L.~Carlone, and J.~A. Castellanos, ``{A Survey on Active Simultaneous Localization and Mapping: State of the Art and New Frontiers},'' \emph{IEEE Trans.~on Robotics (TRO)}, 2023.

\bibitem[Yu et~al.(2021)Yu, Ye, Tancik, and Kanazawa]{yu2020pixelnerf}
A.~Yu, V.~Ye, M.~Tancik, and A.~Kanazawa, ``{pixelNeRF}: Neural radiance fields from one or few images,'' in \emph{Proc.~of the IEEE/CVF Conf.~on Computer Vision and Pattern Recognition (CVPR)}, 2021.

\bibitem[Westheider et~al.(2023)Westheider, R{\"u}ckin, and Popovi{\'c}]{westheider2023multi}
J.~Westheider, J.~R{\"u}ckin, and M.~Popovi{\'c}, ``{Multi-UAV Adaptive Path Planning Using Deep Reinforcement Learning},'' in \emph{Proc.~of the IEEE/RSJ Intl.~Conf.~on Intelligent Robots and Systems (IROS)}, 2023.

\bibitem[Bai et~al.(2017)Bai, Chen, and Englot]{bai2017toward}
S.~Bai, F.~Chen, and B.~Englot, ``{Toward autonomous mapping and exploration for mobile robots through deep supervised learning},'' in \emph{Proc.~of the IEEE/RSJ Intl.~Conf.~on Intelligent Robots and Systems (IROS)}, 2017, pp. 2379--2384.

\bibitem[Chen et~al.(2020)Chen, Martin, Huang, Wang, and Englot]{chen2020autonomous}
F.~Chen, J.~D. Martin, Y.~Huang, J.~Wang, and B.~Englot, ``{Autonomous Exploration Under Uncertainty via Deep Reinforcement Learning on Graphs},'' in \emph{Proc.~of the IEEE/RSJ Intl.~Conf.~on Intelligent Robots and Systems (IROS)}, 2020.

\bibitem[Sukkar et~al.(2019)Sukkar, Best, Yoo, and Fitch]{sukkar2019multi}
F.~Sukkar, G.~Best, C.~Yoo, and R.~Fitch, ``Multi-robot region-of-interest reconstruction with dec-mcts,'' in \emph{Proc.~of the IEEE Intl.~Conf.~on Robotics \& Automation (ICRA)}, 2019.

\bibitem[Choudhury et~al.(2018)Choudhury, Bhardwaj, Kapoor, Ranade, Scherer, and Dey]{choudhury2018data}
S.~Choudhury, M.~Bhardwaj, A.~Kapoor, G.~Ranade, S.~Scherer, and D.~Dey, ``{Data-driven planning via imitation learning},'' \emph{Intl.~Journal~of Robotics Research (IJRR)}, vol.~37, pp. 1632--1672, 2018.

\bibitem[Lodel et~al.(2022)Lodel, Brito, Serra-G\'{o}mez, Ferranti, Babu\v{s}ka, and Alonso-Mora]{lodel2022where}
M.~Lodel, B.~Brito, A.~Serra-G\'{o}mez, L.~Ferranti, R.~Babu\v{s}ka, and J.~Alonso-Mora, ``{Where to Look Next: Learning Viewpoint Recommendations for Informative Trajectory Planning},'' in \emph{Proc.~of the IEEE Intl.~Conf.~on Robotics \& Automation (ICRA)}, 2022.

\bibitem[Chen et~al.(2019)Chen, Bai, Shan, and Englot]{chen2019self}
F.~Chen, S.~Bai, T.~Shan, and B.~Englot, ``{Self-Learning Exploration and Mapping for Mobile Robots via Deep Reinforcement Learning},'' in \emph{AIAA SciTech Forum}, 2019.

\bibitem[Chen et~al.(2021)Chen, Szenher, Huang, Wang, Shan, Bai, and Englot]{chen2021zeroshot}
F.~Chen, P.~Szenher, Y.~Huang, J.~Wang, T.~Shan, S.~Bai, and B.~Englot, ``{Zero-Shot Reinforcement Learning on Graphs for Autonomous Exploration Under Uncertainty},'' in \emph{Proc.~of the IEEE Intl.~Conf.~on Robotics \& Automation (ICRA)}, 2021.

\bibitem[Viseras et~al.(2021)Viseras, Meissner, and Marchal]{viseras2021wildfire}
A.~Viseras, M.~Meissner, and J.~Marchal, ``{Wildfire Front Monitoring with Multiple UAVs using Deep Q-Learning},'' \emph{IEEE Access}, 2021.

\bibitem[Georgakis et~al.(2022{\natexlab{a}})Georgakis, Bucher, Arapin, Schmeckpeper, Matni, and Daniilidis]{georgakis2022upen}
G.~Georgakis, B.~Bucher, A.~Arapin, K.~Schmeckpeper, N.~Matni, and K.~Daniilidis, ``{Uncertainty-driven Planner for Exploration and Navigation},'' in \emph{Proc.~of the IEEE Intl.~Conf.~on Robotics \& Automation (ICRA)}, 2022.

\bibitem[Ramakrishnan et~al.(2020)Ramakrishnan, Al-Halah, and Grauman]{ramakrishnan2020occupancy}
S.~K. Ramakrishnan, Z.~Al-Halah, and K.~Grauman, ``{Occupancy Anticipation for Efficient Exploration and Navigation},'' in \emph{Proc.~of the Europ.~Conf.~on Computer Vision (ECCV)}, 2020.

\bibitem[Schmid et~al.(2023)Schmid, Cheema, Reijgwart, Siegwart, Tombari, and Cadena]{schmid2022scexplorer}
L.~Schmid, M.~N. Cheema, V.~Reijgwart, R.~Siegwart, F.~Tombari, and C.~Cadena, ``{SC-Explorer: Incremental 3D Scene Completion for Safe and Efficient Exploration Mapping and Planning},'' \emph{arXiv preprint arXiv:2208.08307}, 2023.

\bibitem[Schmid et~al.(2022)Schmid, Ni, Zhong, Siegwart, and Andersson]{schmid2022fast}
L.~Schmid, C.~Ni, Y.~Zhong, R.~Siegwart, and O.~Andersson, ``{Fast and Compute-Efficient Sampling-Based Local Exploration Planning via Distribution Learning},'' \emph{IEEE Robotics and Automation Letters (RA-L)}, vol.~7, no.~3, pp. 7810--7817, 2022.

\bibitem[Niroui et~al.(2019)Niroui, Zhang, Kashino, and Nejat]{niroui2019deep}
F.~Niroui, K.~Zhang, Z.~Kashino, and G.~Nejat, ``{Deep Reinforcement Learning Robot for Search and Rescue Applications: Exploration in Unknown Cluttered Environments},'' \emph{IEEE Robotics and Automation Letters (RA-L)}, vol.~4, no.~2, pp. 610--617, 2019.

\bibitem[Cao et~al.(2023{\natexlab{b}})Cao, Hou, Wang, Yi, and Sartoretti]{cao2023adriadne}
Y.~Cao, T.~Hou, Y.~Wang, X.~Yi, and S.~Sartoretti, ``{ARiADNE: A Reinforcement learning approach using Attention-based Deep Networks for Exploration},'' \emph{arXiv preprint arXiv:2301.11575}, 2023.

\bibitem[Reinhart et~al.(2020)Reinhart, Dang, Hand, Papachristos, and Alexis]{reinhart2020learning}
R.~Reinhart, T.~Dang, E.~Hand, C.~Papachristos, and K.~Alexis, ``{Learning-based Path Planning for Autonomous Exploration of Subterranean Environments},'' in \emph{Proc.~of the IEEE Intl.~Conf.~on Robotics \& Automation (ICRA)}, 2020.

\bibitem[Zeng et~al.(2022)Zeng, Zaenker, and Bennewitz]{zeng2022deep}
X.~Zeng, T.~Zaenker, and M.~Bennewitz, ``{Deep Reinforcement Learning for Next-Best-View Planning in Agricultural Applications},'' in \emph{Proc.~of the IEEE Intl.~Conf.~on Robotics \& Automation (ICRA)}, 2022.

\bibitem[Shrestha et~al.(2019)Shrestha, Tian, Feng, Tan, and Vaughan]{shrestha2019learned}
R.~Shrestha, F.-P. Tian, W.~Feng, P.~Tan, and R.~Vaughan, ``{Learned Map Prediction for Enhanced Mobile Robot Exploration},'' in \emph{Proc.~of the IEEE Intl.~Conf.~on Robotics \& Automation (ICRA)}, 2019.

\bibitem[Zacchini et~al.(2023)Zacchini, Ridolfi, and Allotta]{zacchini2023informed}
L.~Zacchini, A.~Ridolfi, and B.~Allotta, ``Informed expansion for informative path planning via online distribution learning,'' \emph{Journal on Robotics and Autonomous Systems (RAS)}, vol. 166, p. 104449, 2023.

\bibitem[Liu et~al.(2023)Liu, Deshpande, Qi, Zhao, Madhivanan, and Sen]{liu2023learning}
Z.~Liu, M.~Deshpande, X.~Qi, D.~Zhao, R.~Madhivanan, and A.~Sen, ``{Learning to Explore (L2E): Deep Reinforcement Learning-based Autonomous Exploration for Household Robot},'' in \emph{{Robotics: Science and Systems Workshop on Robot Representations for Scene Understanding, Reasoning, and Planning}}, 2023.

\bibitem[Vasquez-Gomez et~al.(2021)Vasquez-Gomez, Troncoso, Becerra, Sucar, and Murrieta-Cid]{vasquez2021next}
J.~I. Vasquez-Gomez, D.~Troncoso, I.~Becerra, E.~Sucar, and R.~Murrieta-Cid, ``{Next-best-view regression using a 3D convolutional neural network},'' \emph{Machine Vision and Applications}, vol.~32, no.~42, 2021.

\bibitem[Mendoza et~al.(2020)Mendoza, Vasquez-Gomez, Taud, Sucar, and Reta]{mendoza2020supervised}
M.~Mendoza, J.~I. Vasquez-Gomez, H.~Taud, L.~E. Sucar, and C.~Reta, ``Supervised learning of the next-best-view for 3d object reconstruction,'' \emph{Pattern Recognition Letters}, vol. 133, pp. 224--231, 2020.

\bibitem[Caley et~al.(2019)Caley, Lawrance, and Hollinger]{caley2019deep}
J.~A. Caley, N.~R.~J. Lawrance, and G.~A. Hollinger, ``{Deep learning of structured environments for robot search},'' \emph{Autonomous Robots}, vol.~43, pp. 1695--1714, 2019.

\bibitem[Chen et~al.(2022)Chen, Ji, Lin, Hu, Huang, Li, Tan, and Gan]{chen2022learning}
P.~Chen, D.~Ji, K.~Lin, W.~Hu, W.~Huang, T.~Li, M.~Tan, and C.~Gan, ``{Learning Active Camera for Multi-Object Navigation},'' in \emph{Proc.~of the Advances in Neural Information Processing Systems (NIPS)}, 2022, pp. 28\,670--28\,682.

\bibitem[Tao et~al.(2023)Tao, Wu, Li, Cladera, Zhou, Thakur, and Kumar]{tao2023seer}
Y.~Tao, Y.~Wu, B.~Li, F.~Cladera, A.~Zhou, D.~Thakur, and V.~Kumar, ``{SEER: Safe Efficient Exploration for Aerial Robots using Learning to Predict Information Gain},'' in \emph{Proc.~of the IEEE Intl.~Conf.~on Robotics \& Automation (ICRA)}, 2023.

\bibitem[Li et~al.(2023)Li, Debnath, Stein, and Kosecka]{li2023learning}
Y.~Li, A.~Debnath, G.~Stein, and J.~Kosecka, ``{Learning-Augmented Model-Based Planning for Visual Exploration},'' in \emph{Proc.~of the IEEE/RSJ Intl.~Conf.~on Intelligent Robots and Systems (IROS)}, 2023.

\bibitem[Gao and Zhang(2022)]{gao2022cooperative}
M.~Gao and X.~Zhang, ``{Cooperative Search Method for Multiple {UAV}s Based on Deep Reinforcement Learning},'' \emph{Sensors}, vol.~22, no.~18, p. 6737, 2022.

\bibitem[Zwecher et~al.(2022)Zwecher, Iceland, Levy, Hayoun, Gal, and Barel]{zwecher2022integrating}
E.~Zwecher, E.~Iceland, S.~R. Levy, S.~Y. Hayoun, O.~Gal, and A.~Barel, ``{Integrating Deep Reinforcement and Supervised Learning to Expedite Indoor Mapping},'' in \emph{Proc.~of the IEEE Intl.~Conf.~on Robotics \& Automation (ICRA)}, 2022.

\bibitem[Dai et~al.(2022)Dai, Meng, Jin, and Liu]{dai2022camera}
X.-Y. Dai, Q.-H. Meng, S.~Jin, and Y.-B. Liu, ``Camera view planning based on generative adversarial imitation learning in indoor active exploration,'' \emph{{Applied Soft Computing}}, vol. 129, p. 109621, 2022.

\bibitem[Hepp et~al.(2018)Hepp, Dey, Sinha, Kapoor, Joshi, and Hilliges]{hepp2018learn}
B.~Hepp, D.~Dey, S.~N. Sinha, A.~Kapoor, N.~Joshi, and O.~Hilliges, ``{Learn-to-Score: Efficient 3D Scene Exploration by Predicting View Utility},'' in \emph{Proc.~of the Europ.~Conf.~on Computer Vision (ECCV)}, 2018.

\bibitem[Dhami et~al.(2023)Dhami, Sharma, and Tokekar]{dhami2023pred}
H.~Dhami, V.~D. Sharma, and P.~Tokekar, ``{Pred-NBV: Prediction-guided Next-Best-View Planning for 3D Object Reconstruction},'' \emph{arXiv preprint arXiv:2304.11465}, 2023.

\bibitem[Pan et~al.(2023{\natexlab{a}})Pan, Hu, Wei, Dengler, Zaenker, and Bennewitz]{pan2023oneshot}
S.~Pan, H.~Hu, H.~Wei, N.~Dengler, T.~Zaenker, and M.~Bennewitz, ``{One-Shot View Planning for Fast and Complete Unknown Object Reconstruction},'' \emph{arXiv preprint arXiv:2304.00910}, 2023.

\bibitem[Denniston et~al.(2023{\natexlab{b}})Denniston, Salhotra, Kangaslahti, Caron, and Sukhatme]{denniston2023learned}
C.~Denniston, G.~Salhotra, A.~Kangaslahti, D.~A. Caron, and G.~S. Sukhatme, ``{Learned Parameter Selection for Robotic Information Gathering},'' \emph{arXiv preprint arXiv:2303.05022}, 2023.

\bibitem[Song et~al.(2015)Song, Rodriguez, and Teodorescu]{song2015trajectory}
S.~Song, A.~Rodriguez, and M.~Teodorescu, ``Trajectory planning for autonomous nonholonomic vehicles for optimal monitoring of spatial phenomena,'' in \emph{Proc.~of the Intl.~Conf.~on Unmanned Aircraft Systems (ICUAS)}.\hskip 1em plus 0.5em minus 0.4em\relax IEEE, 2015, pp. 40--49.

\bibitem[Popovi{\'c} et~al.(2020)Popovi{\'c}, Vidal-Calleja, Chung, Nieto, and Siegwart]{popovic2020informativelocunc}
M.~Popovi{\'c}, T.~Vidal-Calleja, J.~J. Chung, J.~Nieto, and R.~Siegwart, ``{Informative Path Planning for Active Field Mapping under Localization Uncertainty},'' in \emph{Proc.~of the IEEE Intl.~Conf.~on Robotics \& Automation (ICRA)}, 2020.

\bibitem[Ott et~al.(2023)Ott, Balaban, and Kochenderfer]{ott2023sboaippms}
J.~Ott, E.~Balaban, and M.~J. Kochenderfer, ``Sequential {B}ayesian optimization for adaptive informative path planning with multimodal sensing,'' in \emph{Proc.~of the IEEE Intl.~Conf.~on Robotics \& Automation (ICRA)}, 2023.

\bibitem[Yang et~al.(2023{\natexlab{a}})Yang, Cao, and Sartoretti]{yang2023intent}
T.~Yang, Y.~Cao, and S.~Sartoretti, ``{Intent-based Deep Reinforcement Learning for Multi-agent Informative Path Planning},'' \emph{arXiv preprint arXiv:2303.05351}, 2023.

\bibitem[Yanes~Luis et~al.(2023)Yanes~Luis, Perales~Esteve, Guti{\'e}rrez~Reina, and Toral~Mar{\'i}n]{yanes2023deep}
S.~Yanes~Luis, M.~Perales~Esteve, D.~Guti{\'e}rrez~Reina, and S.~Toral~Mar{\'i}n, \emph{{Deep Reinforcement Learning Applied to Multi-agent Informative Path Planning in Environmental Missions}}, 2023, pp. 31--61.

\bibitem[Hitz et~al.(2017)Hitz, Galceran, Garneau, Pomerleau, and Siegwart]{hitz2017adaptive}
G.~Hitz, E.~Galceran, M.-E. Garneau, F.~Pomerleau, and R.~Siegwart, ``Adaptive continuous-space informative path planning for online environmental monitoring,'' \emph{Journal of Field Robotics (JFR)}, vol.~34, no.~8, pp. 1427--1449, 2017.

\bibitem[Wei and Zheng(2020)]{wei2020informative}
Y.~Wei and R.~Zheng, ``Informative path planning for mobile sensing with reinforcement learning,'' in \emph{IEEE Conference on Computer Communications}, 2020, pp. 864--873.

\bibitem[Vivaldini et~al.(2019)Vivaldini, Martinelli, Guizilini, Souza, Oliviera, Ramos, and Wolf]{vivaldini2019uav}
K.~C.~T. Vivaldini, T.~H. Martinelli, V.~C. Guizilini, J.~R. Souza, M.~D. Oliviera, F.~T. Ramos, and D.~F. Wolf, ``{UAV route planning for active disease classification},'' \emph{Autonomous Robots}, vol.~43, pp. 1137--1153, 2019.

\bibitem[Hollinger and Sukhatme(2014)]{hollinger2014sampling}
G.~A. Hollinger and G.~S. Sukhatme, ``Sampling-based robotic information gathering algorithms,'' \emph{Intl.~Journal~of Robotics Research (IJRR)}, vol.~33, no.~9, pp. 1271--1287, 2014.

\bibitem[Choi and Cielniak(2021)]{choi2021adaptive}
T.~Choi and G.~Cielniak, ``{Adaptive Selection of Informative Path Planning Strategies via Reinforcement Learning},'' in \emph{Proc.~of the Europ.~Conf.~on Mobile Robotics (ECMR)}, 2021.

\bibitem[Cao et~al.(2023{\natexlab{c}})Cao, Hou, Wang, Yi, and Sartoretti]{wang2023spatiotemporal}
Y.~Cao, T.~Hou, Y.~Wang, X.~Yi, and S.~Sartoretti, ``{Spatio-Temporal Attention Network for Persistent Monitoring of Multiple Mobile Targets},'' in \emph{Proc.~of the IEEE/RSJ Intl.~Conf.~on Intelligent Robots and Systems (IROS)}, 2023.

\bibitem[Marchant et~al.(2014)Marchant, Ramos, and Sanner]{marchant2014sequential}
R.~Marchant, F.~Ramos, and S.~Sanner, ``{Sequential Bayesian Optimisation for Spatial-Temporal Monitoring},'' in \emph{Proc.~of the Conf.~on Uncertainty in Artificial Intelligence (UAI)}, 2014, pp. 553--562.

\bibitem[H{\"u}ttenrauch et~al.(2019)H{\"u}ttenrauch, Adrian, Neumann, et~al.]{huttenrauch2019deep}
M.~H{\"u}ttenrauch, S.~Adrian, G.~Neumann \emph{et~al.}, ``Deep reinforcement learning for swarm systems,'' \emph{Journal of Machine Learning Research}, vol.~20, no.~54, pp. 1--31, 2019.

\bibitem[Duecker et~al.(2021)Duecker, Mersch, Hochdahl, and Kreuzer]{duecker2021embedded}
D.~A. Duecker, B.~Mersch, R.~C. Hochdahl, and E.~Kreuzer, ``{Embedded Stochastic Field Exploration with Micro Diving Agents using {B}ayesian Optimization-Guided Tree-Search and {GMRF}s},'' in \emph{Proc.~of the IEEE/RSJ Intl.~Conf.~on Intelligent Robots and Systems (IROS)}, 2021.

\bibitem[Tzes et~al.(2023)Tzes, Bousias, Chatzipantazis, and Pappas]{tzes2023graph}
M.~Tzes, N.~Bousias, E.~Chatzipantazis, and G.~J. Pappas, ``{Graph Neural Networks for Multi-Robot Active Information Acquisition},'' in \emph{Proc.~of the IEEE Intl.~Conf.~on Robotics \& Automation (ICRA)}, 2023.

\bibitem[Best et~al.(2019)Best, Cliff, Patten, Mettu, and Fitch]{best2019dec}
G.~Best, O.~Cliff, T.~Patten, R.~R. Mettu, and R.~Fitch, ``Dec-mcts: Decentralized planning for multi-robot active perception,'' \emph{Intl.~Journal~of Robotics Research (IJRR)}, vol.~38, no. 2-3, pp. 316--337, 2019.

\bibitem[Meliou et~al.(2007)Meliou, Krause, Guestrin, and Hellerstein]{meliou2007nonmyopic}
A.~Meliou, A.~Krause, C.~Guestrin, and J.~M. Hellerstein, ``Nonmyopic informative path planning in spatio-temporal models,'' in \emph{Proc.~of the Conf.~on Advancements of Artificial Intelligence (AAAI)}, vol.~10, no.~4, 2007, pp. 16--7.

\bibitem[Jin et~al.(2023)Jin, Chen, Rückin, and Popović]{jin2023neunbv}
L.~Jin, X.~Chen, J.~Rückin, and M.~Popović, ``{NeU-NBV: Next Best View Planning Using Uncertainty Estimation in Image-Based Neural Rendering},'' in \emph{Proc.~of the IEEE/RSJ Intl.~Conf.~on Intelligent Robots and Systems (IROS)}, 2023.

\bibitem[Pan et~al.(2022)Pan, Lai, Song, and Huang]{pan2022activenerf}
X.~Pan, Z.~Lai, S.~Song, and G.~Huang, ``{ActiveNeRF: Learning Where to See with Uncertainty Estimation},'' in \emph{Proc.~of the Europ.~Conf.~on Computer Vision (ECCV)}, 2022.

\bibitem[Ran et~al.(2022)Ran, Zeng, He, Li, Chen, Lee, Chen, and Ye]{ran2022neurar}
Y.~Ran, J.~Zeng, S.~He, L.~Li, Y.~Chen, G.~Lee, J.~Chen, and Q.~Ye, ``{NeurAR: Neural Uncertainty for Autonomous 3D Reconstruction with Implicit Neural Representations},'' \emph{IEEE Robotics and Automation Letters (RA-L)}, vol.~8, no.~2, pp. 1125--1132, 2022.

\bibitem[Zhan et~al.(2023)Zhan, Zheng, Xi, Reid, and Rezatofighi]{zhan2022activermap}
H.~Zhan, J.~Zheng, Y.~Xi, I.~Reid, and H.~Rezatofighi, ``{ActiveRMAP: Radiance Field for Active Mapping And Planning},'' \emph{arXiv preprint arXiv:2211.12656}, 2023.

\bibitem[Sünderhauf et~al.(2023)Sünderhauf, Abou-Chakra, and Miller]{sunderhauf2023density}
N.~Sünderhauf, J.~Abou-Chakra, and D.~Miller, ``{Density-aware NeRF Ensembles: Quantifying Predictive Uncertainty in Neural Radiance Fields},'' in \emph{Proc.~of the IEEE Intl.~Conf.~on Robotics \& Automation (ICRA)}, 2023.

\bibitem[Pan et~al.(2023{\natexlab{b}})Pan, Jin, Hu, Popović, and Bennewitz]{pan2023views}
S.~Pan, L.~Jin, H.~Hu, M.~Popović, and M.~Bennewitz, ``{How Many Views Are Needed to Reconstruct an Unknown Object Using NeRF?}'' \emph{arXiv preprint arXiv:2310.00684}, 2023.

\bibitem[Yan et~al.(2023)Yan, Liu, Quan, Chen, and Fu]{yan2023active}
D.~Yan, J.~Liu, F.~Quan, H.~Chen, and M.~Fu, ``{Active Implicit Object Reconstruction Using Uncertainty-Guided Next-Best-View Optimization},'' \emph{IEEE Robotics and Automation Letters (RA-L)}, vol.~8, no.~10, pp. 6395--6402, 2023.

\bibitem[Lee et~al.(2022)Lee, Chen, Wang, Liniger, Kumar, and Yu]{lee2022uncertainty}
S.~Lee, L.~Chen, J.~Wang, A.~Liniger, S.~Kumar, and F.~Yu, ``{Uncertainty Guided Policy for Active Robotic 3D Reconstruction Using Neural Radiance Fields},'' \emph{IEEE Robotics and Automation Letters (RA-L)}, vol.~7, no.~4, pp. 12\,070--12\,077, 2022.

\bibitem[Everingham et~al.(2010)Everingham, Van~Gool, Williams, Winn, and Zisserman]{everingham2010pascal}
M.~Everingham, L.~Van~Gool, C.~K. Williams, J.~Winn, and A.~Zisserman, ``{The Pascal Visual Object Classes (VOC) Challenge},'' \emph{Intl.~Journal~of Computer Vision (IJCV)}, vol.~88, no.~2, pp. 303--338, 2010.

\bibitem[Cordts et~al.(2016)Cordts, Omran, Ramos, Rehfeld, Enzweiler, Benenson, Franke, Roth, and Schiele]{cordts2016cityscapes}
M.~Cordts, M.~Omran, S.~Ramos, T.~Rehfeld, M.~Enzweiler, R.~Benenson, U.~Franke, S.~Roth, and B.~Schiele, ``{The Cityscapes Dataset for Semantic Urban Scene Understanding},'' in \emph{Proc.~of the IEEE/CVF Conf.~on Computer Vision and Pattern Recognition (CVPR)}, 2016, pp. 3213--3223.

\bibitem[Williams and Rasmussen(2006)]{williams2006gaussian}
C.~K. Williams and C.~E. Rasmussen, \emph{Gaussian processes for machine learning}.\hskip 1em plus 0.5em minus 0.4em\relax MIT press Cambridge, MA, 2006, vol.~2, no.~3.

\bibitem[Blum et~al.(2019)Blum, Rohrbach, Popovic, Bartolomei, and Siegwart]{blum2019active}
H.~Blum, S.~Rohrbach, M.~Popovic, L.~Bartolomei, and R.~Siegwart, ``Active learning for uav-based semantic mapping,'' \emph{arXiv preprint arXiv:1908.11157}, 2019.

\bibitem[Zurbrügg et~al.(2022)Zurbrügg, Blum, Cadena, Siegwart, and Schmid]{zurbrugg2022embodied}
R.~Zurbrügg, H.~Blum, C.~Cadena, R.~Siegwart, and L.~Schmid, ``{Embodied Active Domain Adaptation for Semantic Segmentation via Informative Path Planning},'' \emph{IEEE Robotics and Automation Letters (RA-L)}, vol.~7, no.~4, pp. 8691--8698, 2022.

\bibitem[R{\"u}ckin et~al.(2023)R{\"u}ckin, Magistri, Stachniss, and Popović]{ruckin2023informative}
J.~R{\"u}ckin, F.~Magistri, C.~Stachniss, and M.~Popović, ``{An Informative Path Planning Framework for Active Learning in UAV-Based Semantic Mapping},'' \emph{IEEE Trans.~on Robotics (TRO)}, vol.~39, no.~6, pp. 4279--4296, 2023.

\bibitem[R{\"u}ckin et~al.(2024)R{\"u}ckin, Magistri, Stachniss, and Popovi{\'c}]{ruckin2023semi}
J.~R{\"u}ckin, F.~Magistri, C.~Stachniss, and M.~Popovi{\'c}, ``{Semi-Supervised Active Learning for Semantic Segmentation in Unknown Environments Using Informative Path Planning},'' \emph{IEEE Robotics and Automation Letters (RA-L)}, pp. 1--8, 2024.

\bibitem[Chaplot et~al.(2021)Chaplot, Dalal, Gupta, Malik, and Salakhutdinov]{chaplot2021seal}
D.~S. Chaplot, M.~Dalal, S.~Gupta, J.~Malik, and R.~R. Salakhutdinov, ``Seal: Self-supervised embodied active learning using exploration and 3d consistency,'' \emph{Proc.~of the Advances in Neural Information Processing Systems (NIPS)}, vol.~34, pp. 13\,086--13\,098, 2021.

\bibitem[Gazani et~al.(2023)Gazani, Tucsok, Mantegh, and Najjaran]{gazani2023bag}
S.~H. Gazani, M.~Tucsok, I.~Mantegh, and H.~Najjaran, ``{Bag of Views: An Appearance-based Approach to Next-Best-View Planning for 3D Reconstruction},'' \emph{arXiv preprint arXiv:2307.05832}, 2023.

\bibitem[Georgakis et~al.(2022{\natexlab{b}})Georgakis, Bucher, Schmeckpeper, Singh, and Daniilidis]{georgakis2022learning}
G.~Georgakis, B.~Bucher, K.~Schmeckpeper, S.~Singh, and K.~Daniilidis, ``{Learning to Map for Active Semantic Goal Navigation},'' in \emph{Proc.~of the Int.~Conf.~on Learning Representations (ICLR)}, 2022.

\bibitem[Velasco et~al.(2020)Velasco, Valente, and Mersha]{velasco2020adaptive}
O.~Velasco, J.~Valente, and A.~Y. Mersha, ``An adaptive informative path planning algorithm for real-time air quality monitoring using {UAV}s,'' in \emph{Proc.~of the Intl.~Conf.~on Unmanned Aircraft Systems (ICUAS)}.\hskip 1em plus 0.5em minus 0.4em\relax IEEE, 2020, pp. 1121--1130.

\bibitem[Saroya et~al.(2020)Saroya, Best, and Hollinger]{saroya2020online}
M.~Saroya, G.~Best, and G.~A. Hollinger, ``{Online Exploration of Tunnel Networks Leveraging Topological CNN-based World Predictions},'' in \emph{Proc.~of the IEEE/RSJ Intl.~Conf.~on Intelligent Robots and Systems (IROS)}, 2020.

\bibitem[Zhang et~al.(2023)Zhang, Wang, Han, Li, Zhao, Wang, Duan, Fang, Li, and He]{zhang2023affordance}
X.~Zhang, D.~Wang, S.~Han, W.~Li, B.~Zhao, Z.~Wang, X.~Duan, C.~Fang, X.~Li, and J.~He, ``{Affordance-Driven Next-Best-View Planning for Robotic Grasping},'' in \emph{Proc.~of the Conf.~on Robot Learning (CoRL)}, 2023.

\bibitem[Kumar et~al.(2022)Kumar, Essa, and Ha]{kumar2022graph}
K.~N. Kumar, I.~Essa, and S.~Ha, ``Graph-based cluttered scene generation and interactive exploration using deep reinforcement learning,'' in \emph{Proc.~of the IEEE Intl.~Conf.~on Robotics \& Automation (ICRA)}, 2022.

\bibitem[Binney and Sukhatme(2012)]{binney2012branch}
J.~Binney and G.~S. Sukhatme, ``{Branch and bound for informative path planning},'' in \emph{Proc.~of the IEEE Intl.~Conf.~on Robotics \& Automation (ICRA)}.\hskip 1em plus 0.5em minus 0.4em\relax IEEE, 2012, pp. 2147--2154.

\bibitem[Karaman and Frazzoli(2011)]{karaman2011sampling}
S.~Karaman and E.~Frazzoli, ``Sampling-based algorithms for optimal motion planning,'' \emph{Intl.~Journal~of Robotics Research (IJRR)}, vol.~30, no.~7, pp. 846--894, 2011.

\bibitem[Yamauchi(1997)]{yamauchi1997frontier}
B.~Yamauchi, ``A frontier-based approach for autonomous exploration,'' in \emph{Proc.~of the IEEE Intl.~Conf.~on Robotics \& Automation (ICRA)}, 1997, pp. 146--151.

\bibitem[Gelbart et~al.(2014)Gelbart, Snoek, and Adams]{gelbart2014bayesian}
M.~A. Gelbart, J.~Snoek, and R.~P. Adams, ``Bayesian optimization with unknown constraints,'' \emph{arXiv preprint arXiv:1403.5607}, 2014.

\bibitem[Hansen and Ostermeier(2001)]{hansen2001completely}
N.~Hansen and A.~Ostermeier, ``Completely derandomized self-adaptation in evolution strategies,'' \emph{Evolutionary computation}, vol.~9, no.~2, pp. 159--195, 2001.

\bibitem[Lim et~al.(2016)Lim, Hsu, and Lee]{lim2016adaptive}
Z.~W. Lim, D.~Hsu, and W.~S. Lee, ``Adaptive informative path planning in metric spaces,'' \emph{Intl.~Journal~of Robotics Research (IJRR)}, vol.~35, no.~5, pp. 585--598, 2016.

\bibitem[Isler et~al.(2016)Isler, Sabzevari, Delmerico, and Scaramuzza]{isler2016information}
S.~Isler, R.~Sabzevari, J.~Delmerico, and D.~Scaramuzza, ``{An information gain formulation for active volumetric 3D reconstruction},'' in \emph{Proc.~of the IEEE Intl.~Conf.~on Robotics \& Automation (ICRA)}, 2016.

\bibitem[Li et~al.(2020)Li, Gama, Ribeiro, and Prorok]{li2020graph}
Q.~Li, F.~Gama, A.~Ribeiro, and A.~Prorok, ``{Graph Neural Networks for Decentralized Multi-Robot Path Planning},'' in \emph{Proc.~of the IEEE/RSJ Intl.~Conf.~on Intelligent Robots and Systems (IROS)}, 2020.

\bibitem[Ly and Tsai(2019)]{ly2019autonomous}
L.~Ly and Y.-H.~R. Tsai, ``{Autonomous Exploration, Reconstruction, and Surveillance of 3D Environments Aided by Deep Learning},'' in \emph{Proc.~of the IEEE Intl.~Conf.~on Robotics \& Automation (ICRA)}, 2019.

\bibitem[Hanlon et~al.(2023)Hanlon, Sun, Pollefeys, and Blum]{hanlon2023active}
M.~Hanlon, B.~Sun, M.~Pollefeys, and H.~Blum, ``{Active Visual Localization for Multi-Agent Collaboration: A Data-Driven Approach},'' \emph{arXiv preprint arXiv:2310.02650}, 2023.

\bibitem[Caley and Hollinger(2020)]{caley2020environment}
J.~A. Caley and G.~A. Hollinger, ``{Environment Prediction from Sparse Samples for Robotic Information Gathering},'' in \emph{Proc.~of the IEEE Intl.~Conf.~on Robotics \& Automation (ICRA)}, 2020.

\bibitem[Ott et~al.(2024{\natexlab{a}})Ott, Kochenderfer, and Boyd]{ott2024approximate}
J.~Ott, M.~J. Kochenderfer, and S.~Boyd, ``Approximate sequential optimization for informative path planning,'' \emph{arXiv preprint arXiv:2402.08841}, 2024.

\bibitem[Ott et~al.(2024{\natexlab{b}})Ott, Balaban, and Kochenderfer]{ott2024trajectory}
J.~Ott, E.~Balaban, and M.~Kochenderfer, ``Trajectory optimization for adaptive informative path planning with multimodal sensing,'' \emph{arXiv preprint arXiv:2404.18374}, 2024.

\bibitem[Kochenderfer et~al.(2022)Kochenderfer, Wheeler, and Wray]{kochenderfer2022algorithms}
M.~J. Kochenderfer, T.~A. Wheeler, and K.~H. Wray, \emph{Algorithms for {D}ecision {M}aking}.\hskip 1em plus 0.5em minus 0.4em\relax MIT Press, 2022.

\bibitem[Silver et~al.(2017)Silver, Hubert, Schrittwieser, Antonoglou, Lai, Guez, Lanctot, Sifre, Kumaran, Graepel, et~al.]{silver2017mastering}
D.~Silver, T.~Hubert, J.~Schrittwieser, I.~Antonoglou, M.~Lai, A.~Guez, M.~Lanctot, L.~Sifre, D.~Kumaran, T.~Graepel \emph{et~al.}, ``Mastering chess and shogi by self-play with a general reinforcement learning algorithm,'' \emph{arXiv preprint arXiv:1712.01815}, 2017.

\bibitem[Silver et~al.(2018)Silver, Hubert, Schrittwieser, Antonoglou, Lai, Guez, Lanctot, Sifre, Kumaran, Graepel, et~al.]{silver2018general}
------, ``A general reinforcement learning algorithm that masters chess, shogi, and go through self-play,'' \emph{Science}, vol. 362, no. 6419, pp. 1140--1144, 2018.

\bibitem[Hill et~al.(2018)Hill, Raffin, Ernestus, Gleave, Kanervisto, Traore, Dhariwal, Hesse, Klimov, Nichol, Plappert, Radford, Schulman, Sidor, and Wu]{stable-baselines}
A.~Hill, A.~Raffin, M.~Ernestus, A.~Gleave, A.~Kanervisto, R.~Traore, P.~Dhariwal, C.~Hesse, O.~Klimov, A.~Nichol, M.~Plappert, A.~Radford, J.~Schulman, S.~Sidor, and Y.~Wu, ``Stable baselines,'' \url{https://github.com/hill-a/stable-baselines}, 2018.

\bibitem[DeepMind et~al.(2020)DeepMind, Babuschkin, Baumli, Bell, Bhupatiraju, Bruce, Buchlovsky, Budden, Cai, Clark, Danihelka, Dedieu, Fantacci, Godwin, Jones, Hemsley, Hennigan, Hessel, Hou, Kapturowski, Keck, Kemaev, King, Kunesch, Martens, Merzic, Mikulik, Norman, Papamakarios, Quan, Ring, Ruiz, Sanchez, Sartran, Schneider, Sezener, Spencer, Srinivasan, Stanojevi\'{c}, Stokowiec, Wang, Zhou, and Viola]{deepmind2020jax}
\BIBentryALTinterwordspacing
DeepMind, I.~Babuschkin, K.~Baumli, A.~Bell, S.~Bhupatiraju, J.~Bruce, P.~Buchlovsky, D.~Budden, T.~Cai, A.~Clark, I.~Danihelka, A.~Dedieu, C.~Fantacci, J.~Godwin, C.~Jones, R.~Hemsley, T.~Hennigan, M.~Hessel, S.~Hou, S.~Kapturowski, T.~Keck, I.~Kemaev, M.~King, M.~Kunesch, L.~Martens, H.~Merzic, V.~Mikulik, T.~Norman, G.~Papamakarios, J.~Quan, R.~Ring, F.~Ruiz, A.~Sanchez, L.~Sartran, R.~Schneider, E.~Sezener, S.~Spencer, S.~Srinivasan, M.~Stanojevi\'{c}, W.~Stokowiec, L.~Wang, G.~Zhou, and F.~Viola, ``The {D}eep{M}ind {JAX} {E}cosystem,'' 2020. [Online]. Available: \url{http://github.com/deepmind}
\BIBentrySTDinterwordspacing

\bibitem[Egorov et~al.(2017)Egorov, Sunberg, Balaban, Wheeler, Gupta, and Kochenderfer]{egorov2017pomdps}
\BIBentryALTinterwordspacing
M.~Egorov, Z.~N. Sunberg, E.~Balaban, T.~A. Wheeler, J.~K. Gupta, and M.~J. Kochenderfer, ``{POMDP}s.jl: A framework for sequential decision making under uncertainty,'' \emph{Journal of Machine Learning Research}, vol.~18, no.~26, pp. 1--5, 2017. [Online]. Available: \url{http://jmlr.org/papers/v18/16-300.html}
\BIBentrySTDinterwordspacing

\bibitem[Ansel et~al.(2024)Ansel, Yang, He, Gimelshein, Jain, Voznesensky, Bao, Bell, Berard, Burovski, Chauhan, Chourdia, Constable, Desmaison, DeVito, Ellison, Feng, Gong, Gschwind, Hirsh, Huang, Kalambarkar, Kirsch, Lazos, Lezcano, Liang, Liang, Lu, Luk, Maher, Pan, Puhrsch, Reso, Saroufim, Siraichi, Suk, Suo, Tillet, Wang, Wang, Wen, Zhang, Zhao, Zhou, Zou, Mathews, Chanan, Wu, and Chintala]{Ansel_PyTorch_2_Faster_2024}
\BIBentryALTinterwordspacing
J.~Ansel, E.~Yang, H.~He, N.~Gimelshein, A.~Jain, M.~Voznesensky, B.~Bao, P.~Bell, D.~Berard, E.~Burovski, G.~Chauhan, A.~Chourdia, W.~Constable, A.~Desmaison, Z.~DeVito, E.~Ellison, W.~Feng, J.~Gong, M.~Gschwind, B.~Hirsh, S.~Huang, K.~Kalambarkar, L.~Kirsch, M.~Lazos, M.~Lezcano, Y.~Liang, J.~Liang, Y.~Lu, C.~Luk, B.~Maher, Y.~Pan, C.~Puhrsch, M.~Reso, M.~Saroufim, M.~Y. Siraichi, H.~Suk, M.~Suo, P.~Tillet, E.~Wang, X.~Wang, W.~Wen, S.~Zhang, X.~Zhao, K.~Zhou, R.~Zou, A.~Mathews, G.~Chanan, P.~Wu, and S.~Chintala, ``{PyTorch 2: Faster Machine Learning Through Dynamic Python Bytecode Transformation and Graph Compilation},'' in \emph{29th ACM International Conference on Architectural Support for Programming Languages and Operating Systems, Volume 2 (ASPLOS '24)}.\hskip 1em plus 0.5em minus 0.4em\relax ACM, Apr. 2024. [Online]. Available: \url{https://pytorch.org/assets/pytorch2-2.pdf}
\BIBentrySTDinterwordspacing

\bibitem[Abadi et~al.(2015)Abadi, Agarwal, Barham, Brevdo, Chen, Citro, Corrado, Davis, Dean, Devin, Ghemawat, Goodfellow, Harp, Irving, Isard, Jozefowicz, Jia, Kaiser, Kudlur, Levenberg, Mané, Schuster, Monga, Moore, Murray, Olah, Shlens, Steiner, Sutskever, Talwar, Tucker, Vanhoucke, Vasudevan, Viégas, Vinyals, Warden, Wattenberg, Wicke, Yu, and Zheng]{Abadi_TensorFlow_Large-scale_machine_2015}
M.~Abadi, A.~Agarwal, P.~Barham, E.~Brevdo, Z.~Chen, C.~Citro, G.~S. Corrado, A.~Davis, J.~Dean, M.~Devin, S.~Ghemawat, I.~Goodfellow, A.~Harp, G.~Irving, M.~Isard, R.~Jozefowicz, Y.~Jia, L.~Kaiser, M.~Kudlur, J.~Levenberg, D.~Mané, M.~Schuster, R.~Monga, S.~Moore, D.~Murray, C.~Olah, J.~Shlens, B.~Steiner, I.~Sutskever, K.~Talwar, P.~Tucker, V.~Vanhoucke, V.~Vasudevan, F.~Viégas, O.~Vinyals, P.~Warden, M.~Wattenberg, M.~Wicke, Y.~Yu, and X.~Zheng, ``{TensorFlow, Large-scale machine learning on heterogeneous systems},'' Nov. 2015.

\bibitem[Bayerlein et~al.(2021)Bayerlein, Theile, Caccamo, and Gesbert]{bayerlein2021multi}
H.~Bayerlein, M.~Theile, M.~Caccamo, and D.~Gesbert, ``{Multi-UAV Path Planning for Wireless Data Harvesting With Deep Reinforcement Learning},'' \emph{IEEE Open Journal of the Communications Society}, vol.~2, pp. 1171--1187, 2021.

\bibitem[Yang et~al.(2023{\natexlab{b}})Yang, Liu, Koga, Ashgharivaskasi, and Atanasov]{yang2023learning}
P.~Yang, Y.~Liu, S.~Koga, A.~Ashgharivaskasi, and N.~Atanasov, ``{Learning Continuous Control Policies for Information-Theoretic Active Perception},'' in \emph{Proc.~of the IEEE Intl.~Conf.~on Robotics \& Automation (ICRA)}, 2023.

\bibitem[Bartolomei et~al.(2021)Bartolomei, Teixeira, and Chli]{bartolomei2021semantic}
L.~Bartolomei, L.~Teixeira, and M.~Chli, ``Semantic-aware active perception for {UAV}s using deep reinforcement learning,'' in \emph{Proc.~of the IEEE/RSJ Intl.~Conf.~on Intelligent Robots and Systems (IROS)}, 2021.

\bibitem[Kaelbling et~al.(1998)Kaelbling, Littman, and Cassandra]{kaelbling1998planning}
L.~Kaelbling, M.~L. Littman, and A.~R. Cassandra, ``Planning and acting in partially observable stochastic domains,'' \emph{{Artificial Intelligence}}, vol. 101, no. 1-2, pp. 99--134, 1998.

\bibitem[Qie et~al.(2019)Qie, Shi, Shen, Xu, Li, and Wang]{qie2019joint}
H.~Qie, D.~Shi, T.~Shen, X.~Xu, Y.~Li, and L.~Wang, ``Joint optimization of multi-{UAV} target assignment and path planning based on multi-agent reinforcement learning,'' \emph{IEEE access}, vol.~7, pp. 146\,264--146\,272, 2019.

\bibitem[Fan et~al.(2020)Fan, Long, Liu, and Pan]{fan2020distributed}
T.~Fan, P.~Long, W.~Liu, and J.~Pan, ``Distributed multi-robot collision avoidance via deep reinforcement learning for navigation in complex scenarios,'' \emph{Intl.~Journal~of Robotics Research (IJRR)}, vol.~39, no.~7, pp. 856--892, 2020.

\bibitem[Arora et~al.(2019)Arora, Furlong, Fitch, Sukkarieh, and Fong]{arora2019multi}
A.~Arora, P.~M. Furlong, R.~Fitch, S.~Sukkarieh, and T.~Fong, ``Multi-modal active perception for information gathering in science missions,'' \emph{Autonomous Robots}, vol.~43, pp. 1827--1853, 2019.

\bibitem[Bucher et~al.(2021)Bucher, Schmeckpeper, Matni, and Daniilidis]{bucher2021adversiarial}
B.~Bucher, K.~Schmeckpeper, N.~Matni, and K.~Daniilidis, ``{An Adversarial Objective for Scalable Exploration},'' in \emph{Proc.~of the IEEE/RSJ Intl.~Conf.~on Intelligent Robots and Systems (IROS)}, 2021, pp. 2670--2677.

\bibitem[Ott et~al.(2022)Ott, Kim, Bouman, Peltzer, Sobue, Delecki, Kochenderfer, Burdick, and Agha-mohammadi]{ott2022risk}
J.~Ott, S.-K. Kim, A.~Bouman, O.~Peltzer, M.~Sobue, H.~Delecki, M.~J. Kochenderfer, J.~Burdick, and A.-a. Agha-mohammadi, ``Risk-aware meta-level decision making for exploration under uncertainty,'' \emph{arXiv preprint arXiv:2209.05580}, 2022.

\bibitem[Bouman et~al.(2022)Bouman, Ott, Kim, Chen, Kochenderfer, Lopez, Agha-mohammadi, and Burdick]{bouman2022adaptive}
A.~Bouman, J.~Ott, S.-K. Kim, K.~Chen, M.~J. Kochenderfer, B.~Lopez, A.-a. Agha-mohammadi, and J.~Burdick, ``Adaptive coverage path planning for efficient exploration of unknown environments,'' in \emph{2022 IEEE/RSJ International Conference on Intelligent Robots and Systems (IROS)}.\hskip 1em plus 0.5em minus 0.4em\relax IEEE, 2022, pp. 11\,916--11\,923.

\bibitem[Ye et~al.(2018)Ye, Lin, Li, Zheng, and Yang]{ye2018active}
X.~Ye, Z.~Lin, H.~Li, S.~Zheng, and Y.~Yang, ``{Active Object Perceiver: Recognition-guided Policy Learning for Object Searching on Mobile Robots},'' in \emph{Proc.~of the IEEE/RSJ Intl.~Conf.~on Intelligent Robots and Systems (IROS)}, 2018.

\bibitem[Pomerleau(1991)]{pomerleau1991efficient}
D.~A. Pomerleau, ``Efficient training of artificial neural networks for autonomous navigation,'' \emph{Neural Computation}, vol.~3, no.~1, pp. 88--97, 1991.

\bibitem[Settles(2009)]{settles2009active}
B.~Settles, ``{Active learning literature survey},'' University of Wisconsin-Madison, Department of Computer Sciences, Tech. Rep., 2009.

\bibitem[Kiefer(1959)]{kiefer1959optimum}
J.~Kiefer, ``Optimum experimental designs,'' \emph{Journal of the Royal Statistical Society: Series B (Methodological)}, vol.~21, no.~2, pp. 272--304, 1959.

\bibitem[R{\"u}ckin et~al.(2022{\natexlab{b}})R{\"u}ckin, Jin, Magistri, Stachniss, and Popovi{\'c}]{ruckin2022informative}
J.~R{\"u}ckin, L.~Jin, F.~Magistri, C.~Stachniss, and M.~Popovi{\'c}, ``{Informative Path Planning for Active Learning in Aerial Semantic Mapping},'' in \emph{Proc.~of the IEEE/RSJ Intl.~Conf.~on Intelligent Robots and Systems (IROS)}, 2022.

\bibitem[Abdar et~al.(2021)Abdar, Pourpanah, Hussain, Rezazadegan, Liu, Ghavamzadeh, Fieguth, Cao, Khosravi, Acharya, et~al.]{abdar2021review}
M.~Abdar, F.~Pourpanah, S.~Hussain, D.~Rezazadegan, L.~Liu, M.~Ghavamzadeh, P.~Fieguth, X.~Cao, A.~Khosravi, U.~R. Acharya \emph{et~al.}, ``{A review of uncertainty quantification in deep learning: Techniques, applications and challenges},'' \emph{Information fusion}, vol.~76, pp. 243--297, 2021.

\bibitem[Zhu et~al.(2023)Zhu, Cheng, Zhang, Cui, Zhang, and Liu]{zhu2023multi}
L.~Zhu, J.~Cheng, H.~Zhang, Z.~Cui, W.~Zhang, and Y.~Liu, ``{Multi-robot Collaborative Area Search Based on Deep Reinforcement Learning},'' \emph{arXiv preprint arXiv:2312.01747}, 2023.

\bibitem[Menon et~al.(2023)Menon, Zaenker, Dengler, and Bennewitz]{menon2023nbvsc}
R.~Menon, T.~Zaenker, N.~Dengler, and M.~Bennewitz, ``{NBV-SC: Next Best View Planning based on Shape Completion for Fruit Mapping and Reconstruction},'' in \emph{Proc.~of the IEEE/RSJ Intl.~Conf.~on Intelligent Robots and Systems (IROS)}, 2023.

\bibitem[Cao et~al.(2022)Cao, Zhu, Yang, Xia, Choset, Oh, and Zhang]{cao2022autonomous}
C.~Cao, H.~Zhu, F.~Yang, Y.~Xia, H.~Choset, J.~Oh, and J.~Zhang, ``{Autonomous Exploration Development Environment and the Planning Algorithms},'' in \emph{Proc.~of the IEEE Intl.~Conf.~on Robotics \& Automation (ICRA)}, 2022.

\bibitem[Novkovic et~al.(2020)Novkovic, Pautrat, Furrer, Breyer, Siegwart, and Nieto]{novkovic2020object}
T.~Novkovic, R.~Pautrat, F.~Furrer, M.~Breyer, R.~Siegwart, and J.~Nieto, ``{Object Finding in Cluttered Scenes Using Interactive Perception},'' in \emph{Proc.~of the IEEE Intl.~Conf.~on Robotics \& Automation (ICRA)}, 2020.

\bibitem[Bousmalis et~al.(2018)Bousmalis, Irpan, Wohlhart, Bai, Kelcey, Kalakrishnan, Downs, Ibarz, Pastor, Konolige, et~al.]{bousmalis2018using}
K.~Bousmalis, A.~Irpan, P.~Wohlhart, Y.~Bai, M.~Kelcey, M.~Kalakrishnan, L.~Downs, J.~Ibarz, P.~Pastor, K.~Konolige \emph{et~al.}, ``{Using simulation and domain adaptation to improve efficiency of deep robotic grasping},'' in \emph{Proc.~of the IEEE Intl.~Conf.~on Robotics \& Automation (ICRA)}, 2018.

\bibitem[Zhao et~al.(2020)Zhao, Queralta, and Westerlund]{zhao2020sim}
W.~Zhao, J.~P. Queralta, and T.~Westerlund, ``{Sim-to-real transfer in deep reinforcement learning for robotics: a survey},'' in \emph{2020 IEEE symposium series on computational intelligence (SSCI)}, 2020.

\bibitem[Zhu et~al.(2020)Zhu, Yu, Gupta, Shah, Hartikainen, Singh, Kumar, and Levine]{zhu2020ingredients}
H.~Zhu, J.~Yu, A.~Gupta, D.~Shah, K.~Hartikainen, A.~Singh, V.~Kumar, and S.~Levine, ``{The ingredients of real-world robotic reinforcement learning},'' in \emph{Proc.~of the Int.~Conf.~on Learning Representations (ICLR)}, 2020.

\bibitem[Kirk et~al.(2023)Kirk, Zhang, Grefenstette, and Rockt{\"a}schel]{kirk2023survey}
R.~Kirk, A.~Zhang, E.~Grefenstette, and T.~Rockt{\"a}schel, ``{A survey of zero-shot generalisation in deep reinforcement learning},'' \emph{Journal of Artificial Intelligence Research (JAIR)}, vol.~76, pp. 201--264, 2023.

\bibitem[Gupta et~al.(2018)Gupta, Mendonca, Liu, Abbeel, and Levine]{gupta2018meta}
A.~Gupta, R.~Mendonca, Y.~Liu, P.~Abbeel, and S.~Levine, ``{Meta-reinforcement learning of structured exploration strategies},'' in \emph{Proc.~of the Advances in Neural Information Processing Systems (NIPS)}, 2018.

\bibitem[Sadeghi and Levine(2017)]{sadeghicad2rl}
F.~Sadeghi and S.~Levine, ``{CAD2RL: Real Single-Image Flight Without a Single Real Image},'' in \emph{Proc.~of Robotics: Science and Systems (RSS)}, 2017.

\bibitem[Genc et~al.(2020)Genc, Mallya, Bodapati, Sun, and Tao]{genc2020zeroshot}
S.~Genc, S.~Mallya, S.~Bodapati, T.~Sun, and Y.~Tao, ``{Zero-Shot Reinforcement Learning with Deep Attention Convolutional Neural Networks},'' in \emph{Proc.~of the Advances in Neural Information Processing Systems (NIPS)}, 2020.

\end{thebibliography}

\end{document}